\documentclass[10pt,twocolumn,letterpaper]{article}

\usepackage{cvpr}              
\usepackage[accsupp]{axessibility}
\usepackage{amsfonts}
\usepackage{amsmath}
\usepackage{bbm}
\usepackage{multicol}
\usepackage{multirow}
\usepackage{colortbl}
\usepackage[symbol]{footmisc}

\usepackage[dvipsnames]{xcolor}

\definecolor{cvprblue}{rgb}{0.21,0.49,0.74}
\usepackage[pagebackref,breaklinks,colorlinks,citecolor=cvprblue,hyperfootnotes=false]{hyperref}

\definecolor{instructioncolor}{rgb}{.0, .0, 0.0}
\newcommand{\rvc}[1]{\textcolor{instructioncolor}{#1}}
\definecolor{instructioncolor2}{rgb}{0.0, .0, 0.0}
\newcommand{\rvvc}[1]{\textcolor{instructioncolor2}{#1}}
\definecolor{instructioncolor3}{rgb}{0.0, 0.0, 0.0}
\newcommand{\abc}[1]{\textcolor{instructioncolor3}{#1}}

\renewcommand*{\fnsymbol}[1]{%
  \ensuremath{%
    \ifcase#1\or 
    \dagger\or   
    \ast\or      
    \ddagger\or  
    \mathsection\or 
    \mathparagraph\or 
    \|\or        
    **\or        
    \dagger\dagger\or 
    \ddagger\ddagger 
    \else \@ctrerr  
    \fi}}
\makeatother

\def\paperID{7943} 
\def\confName{CVPR}
\def\confYear{2024}

\title{Contextrast: Contextual Contrastive Learning for Semantic Segmentation}

\author{Changki Sung$^1$, Wanhee Kim$^{2\dagger}$, Jungho An$^{2\dagger}$, Wooju Lee$^1$, Hyungtae Lim$^{1\ast}$, and Hyun Myung$^{1\ast}$\\
$^1$School of Electrical Engineering, KI-Robotics,\\
Korea Advanced Institute of Science and Technology, Republic of Korea\\
$^2$Department of Automotive Engineering, Kookmin University, Republic of Korea\\
{\tt\small $^1${\{cs1032, dnwn24, shapelim\thanks{Corresponding authors: Dr. Hyungtae Lim and Prof. Hyun Myung}, hmyung$^\ast$ \}@kaist.ac.kr}}
\quad
{\tt\small $^2${\{gml78905\thanks{Work done during internship at KAIST}, ajh427$^\dagger$\}@kookmin.ac.kr}}
}

\begin{document}
\maketitle

\begin{abstract}

Despite great improvements in semantic segmentation, challenges \abc{persist} because of the lack of local/global contexts and the relationship between them. In this paper, we propose \textit{Contextrast}, \rvc{a} contrastive learning-based semantic segmentation \rvc{method} that allows \rvc{to capture} local/global contexts and \rvc{comprehend} their relationships. Our proposed method comprises two parts: a) contextual contrastive learning (CCL) and b) boundary-aware negative~(BANE) sampling.
Contextual contrastive learning obtains local/global context from multi-scale feature aggregation and inter/intra-relationship of features for better discrimination capabilities. 
Meanwhile, BANE sampling \rvc{selects embedding features along the boundaries of incorrectly predicted regions to employ them as harder negative samples} on our contrastive learning, \abc{resolving segmentation issues along the boundary region by exploiting fine-grained details.}
We demonstrate that our Contextrast substantially enhances the performance of semantic segmentation networks, outperforming state-of-the-art contrastive learning approaches on diverse public datasets,~\eg Cityscapes, CamVid, PASCAL-C, COCO-Stuff, and ADE20K, without an increase in computational cost during inference.

\end{abstract}    
\section{Introduction}
\label{sec:intro}

\begin{figure}[t!]
	\centering 
	\captionsetup{font=footnotesize}
    \begin{subfigure}[b]{0.15\textwidth}
		\includegraphics[width=0.97\textwidth]{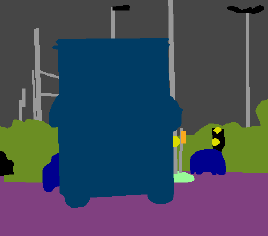}
		\caption{Ground truth}
	\end{subfigure}
    \begin{subfigure}[b]{0.15\textwidth}
		\includegraphics[width=0.97\textwidth]{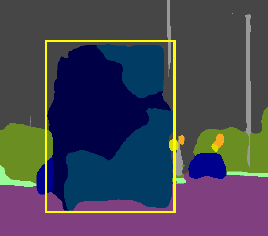}
		\caption{HRNet~\cite{sun2019high}}
	\end{subfigure}
    \begin{subfigure}[b]{0.15\textwidth}
		\includegraphics[width=0.97\textwidth]{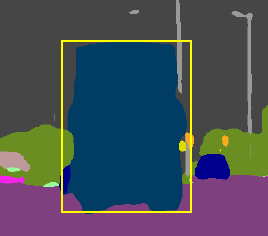}
		\caption{Ours}
	\end{subfigure}
	\begin{subfigure}[b]{0.45\textwidth}
		\includegraphics[width=1.0\textwidth]{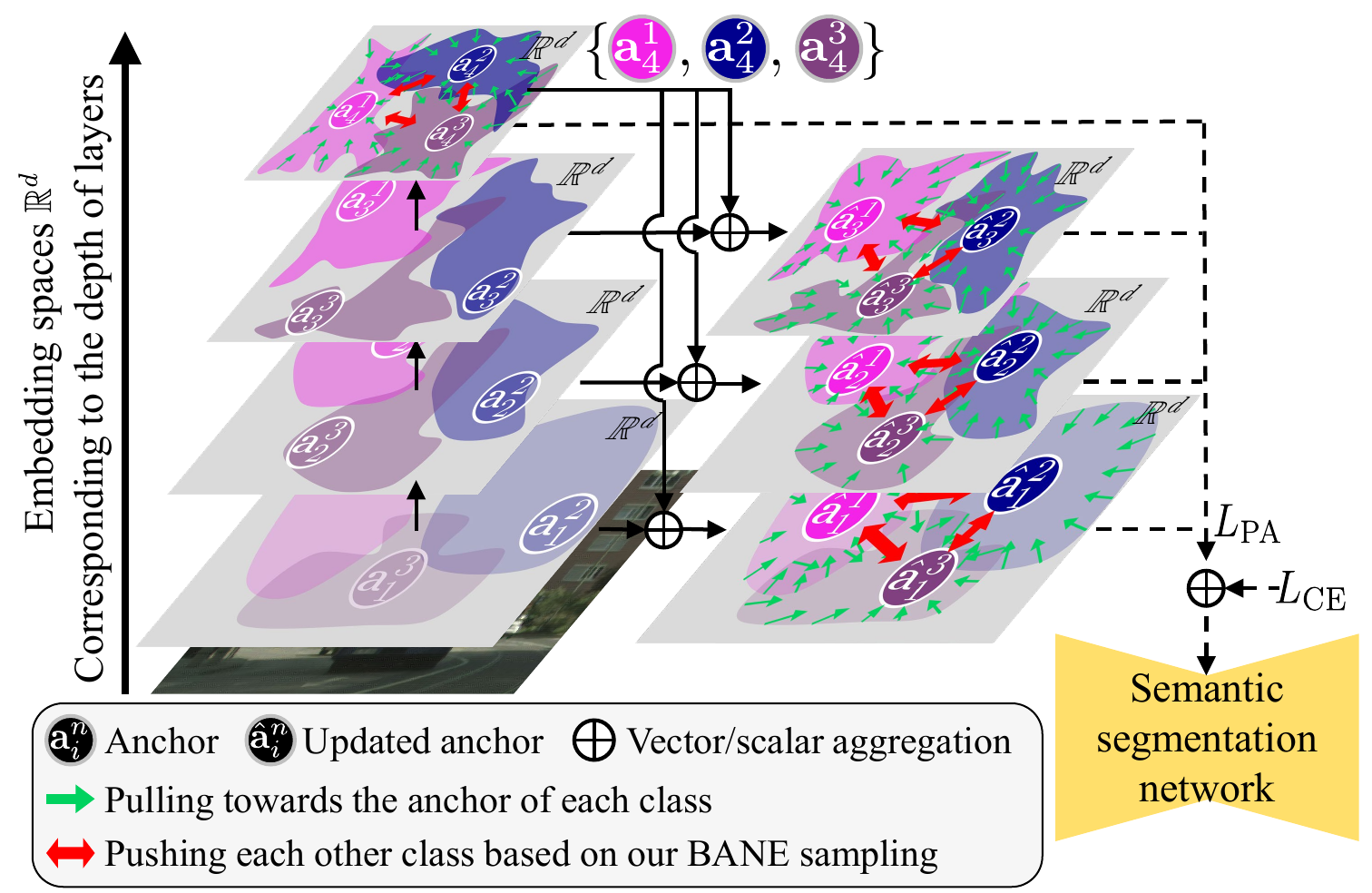}
		\caption{}
	\end{subfigure}
	\begin{subfigure}[b]{0.45\textwidth}
		\includegraphics[width=1.0\textwidth]{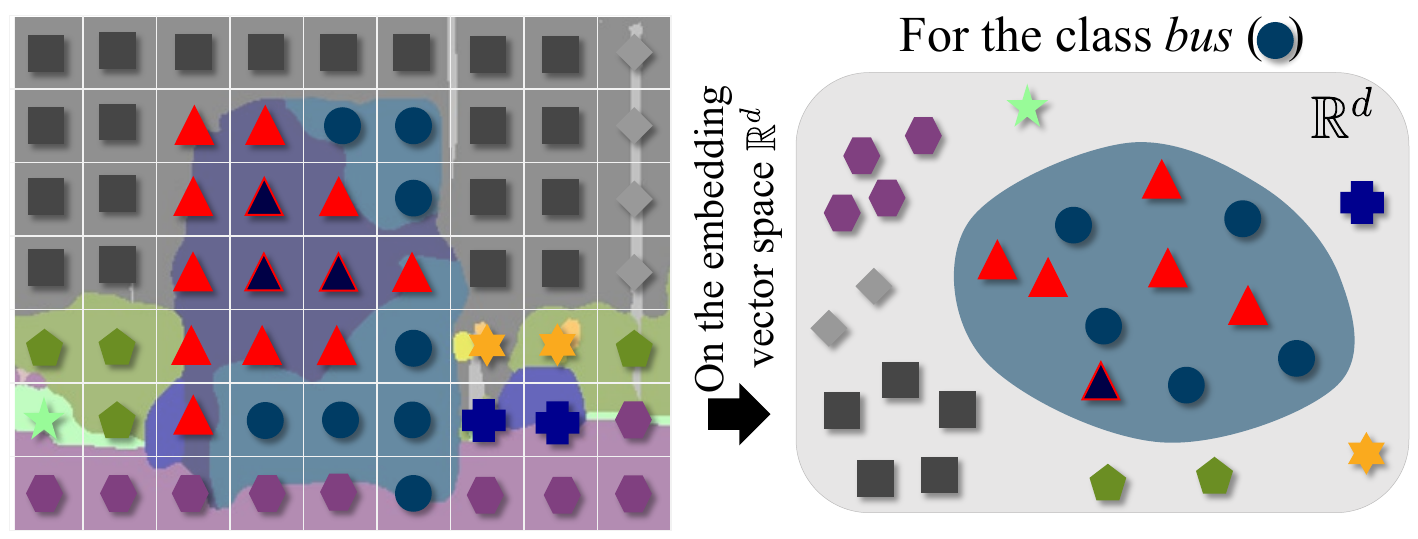}
		\caption{}
	\end{subfigure}	
    \vspace{-0.2cm}
    \caption{(a)~Ground truth, (b)~output of HRNet~\cite{sun2019high}, (c)~and that of ours. (d)~Overview of our contextual contrastive learning (CCL): the representative anchor\rvc{s} of the last layer, which \rvc{are} from the higher embedding space levels, \rvc{are} aggregated to representative anchors of the lower layer to encapsulate local and global context\rvvc{. By doing so, the anchor of the $n$-th class on the $i$-th layer $\mathbf{a}^n_i$ is updated as $\hat{\mathbf{a}}^n_i$~(on the right side, its position is shifted), } 
    \rvvc{enhancing the distinctiveness between anchors of each class}. (e)~Visual description of our boundary-aware negative~(BANE) sampling \rvvc{(triangles with red color and red borders).}
    Our sampling prioritizes selecting the features of incorrect predictions at the edges~(red triangles) rather than those inside the regions (triangles with red borders) as negative samples. Each shape represents an embedding vector derived from the respective class~(best viewed in color).}	
    \label{fig:key_ideas}
    \vspace{-0.4cm}
\end{figure}

Semantic segmentation is a fundamental technique utilized across diverse applications, including autonomous driving, medical imaging, and robotics~\cite{dumitru2023using, sanderson2022fcn, wang2203stepwise, duc2022colonformer, hurtado2022semantic, tzelepi2021semantic}. Recent empirical studies have achieved remarkable advancements in semantic segmentation, benefiting significantly from the availability of extensive datasets~\cite{deng2009imagenet,cordts2016cityscapes,mottaghi2014role,caesar2018coco,brostow2009semantic,zhou2017scene}. \rvvc{To improve segmentation performance, researchers have proposed \rvvc{larger } deep neural network~(DNN) architectures~\cite{chen2014semantic, chen2017deeplab, chen2017rethinking, zhao2017pyramid, chen2018encoder, xiao2018unified, li2018pyramid, ke2018adaptive, yu2018learning, yurtkulu2019semantic, zhou2019fusion, sun2019high, ding2019semantic, fu2019dual, zhang2019dual, yu2020context, yuan2020object, choi2020cars, wang2020deep, hong2021deep, huynh2021progressive, strudel2021segmenter, liu2021swin, li2022deep, xu2023pidnet, zhong2023understanding, wang2023internimage, woo2023convnext} and novel loss functions~\cite{yuan2020segfix,wang2022active,tan2022semantic}.}


\rvvc{Despite these achievements, semantic segmentation sometimes produces inaccurate segmentation, as illustrated in~\cref{fig:key_ideas}(b).}
\rvvc{It could be solved by increasing the complexity of networks~\cite{strudel2021segmenter, liu2021swin, li2022deep, wang2023internimage, woo2023convnext}; however, these approaches require more memory and may potentially slow down the inference speed.
Therefore, it is necessary to improve performance without any additional neural network modules for efficiency.}

\begin{table}[t]
\scriptsize
\centering
\captionsetup{font=footnotesize}
\begin{tabular}{l|ccc}
\hline
\rowcolor[HTML]{DAE8FC} 
                                      & Complexity & Scale        & Boundary awareness \\ \hline
\textit{Baseline}                     & -          & -            & -                  \\
\textit{ICCV 21~\cite{wang2021exploring}}                      & CL, MB     & Single       & -                  \\
\multicolumn{1}{c|}{\textit{ICCV 21~\cite{hu2021region}}} & CL, MB, AL & Single       & -                  \\
\textit{ECCV 22~\cite{pissas2022multi}}                      & CL         & Multi        & -                  \\
\rowcolor[HTML]{EFEFEF} 
\textit{Ours}                         & CL         & Multi fusion & Aware              \\ \hline
\end{tabular}
\caption{\abc{Comparison between SOTAs and ours (CL: contrastive learning, MB: memory bank, AL: additional loss)}.}
\label{tab:prop_diff}
\end{table}

\rvvc{For these reasons}, another noteworthy method, contrastive learning, has emerged as a valuable solution~\cite{wang2021exploring, hu2021region} 
\rvvc{because contrastive learning aims to make the networks understand the semantic context better during the training process.
This is achieved by attaching refinement modules during the training stage and detaching them on the inference so that the network models preserve the inference speed without increasing the complexity of architectures.} 

However, previous \rvc{researches~\cite{wang2021exploring, hu2021region} have} overlooked the significance of multi-scale features, including both global and local contexts.
To mitigate the problem, Pissas~\etal~\cite{pissas2022multi} proposed a method to extract multi-scale embedding features from multiple encoder layers. However, the method could not consistently comprehend the relationships between different scales of features because multi-scale and cross-scale contrastive learning are considered independently. Consequently, it struggles to comprehend relationships between local and global contexts.


To address the aforementioned problems, we propose a supervised contrastive learning framework incorporating two novel methods for semantic segmentation, called \textit{Contextrast}. First, contextual contrastive learning (CCL) is proposed to acquire embedded features from multiple encoder layers representing local and global contexts. Based on embedded features, we define the representative anchors \rvvc{in each layer}, which act as the semantic centroids for each class.
\rvc{The anchors of the last layer represent \rvvc{more} global context than the anchors in the lower layers. The anchors of the last layer are \rvvc{used} to update anchors in each layer~(green arrows in \cref{fig:key_ideas}(d))}. Thus, the anchors in the lower layers can have both global and local contexts.
Consequently, it consistently understands relationships between local and global contexts using the updated representative anchors, which share the same global contexts. 
Second, boundary-aware negative~(BANE) sampling, which is inspired by~\cite{yuan2020segfix} and \cite{wang2022active} that focus on the boundary regions, is proposed to sample negative examples along the boundaries of incorrectly predicted regions~(\cref{fig:key_ideas}(e)). 
It leverages the advantages of sampling harder negative examples and capturing fine-grained details, so the proposed method gets more informative gradients during the training process~\cite{kalantidis2020hard,wang2021exploring}. \abc{We summarize several key properties of state-of-the-art methods and ours, as shown in~\cref{tab:prop_diff}.} 

In sum, this paper makes the following contributions:
\begin{itemize}
\item Our Contextrast enables a segmentation model to capture global/local context information from multi-scale features and consistently comprehend relationships between them through the representative anchors. 
\item Our BANE sampling enables the acquisition of informative negative samples for contrastive learning and fine-grained details. It guides the model to focus on confusion regions progressively during the training.
\item To demonstrate the applicability of our Contextrast in semantic segmentation, we verify the state-of-the-art performance for contrastive learning-based semantic segmentation on various powerful CNN models~\cite{sun2019high,yuan2020object,chen2017rethinking} and public datasets: Cityscapes~\cite{cordts2016cityscapes}, CamVid~\cite{brostow2009semantic}, PASCAL-C~\cite{mottaghi2014role}, COCO-Stuff~\cite{caesar2018coco}, and ADE20K~\cite{zhou2017scene}, which were acquired in different domains.

\end{itemize}

\section{Related works}
\label{sec:related_works}
\subsection{Semantic segmentation}
Semantic segmentation, a fundamental task in computer vision, entails pixel-wise object classification within an image. In recent years, remarkable advancements in deep learning have propelled the field of semantic segmentation to unprecedented levels of accuracy and efficiency. At first, fully connected networks (FCNs)~\cite{long2015fully} brought significant progress in semantic segmentation by introducing end-to-end dense feature learning. 
However, FCNs suffer from limited spatial and contextual information because of the narrow local receptive fields.

Thus, the following researchers focused on capturing better spatial and context information in the semantic segmentation. Atrous spatial pyramid pooling (ASPP)~\cite{chen2017rethinking} captures a diverse range of contextual information. HRNet~\cite{sun2019high} maintains high-resolution representation throughout the network, ensuring the preservation of fine-grained details. For further improvements, OCRNet~\cite{yuan2020object} architecture was introduced that integrates object-contextual representations, allowing the network to consider relationships between objects within a scene. As these advanced methods learn discrimination ability using contextual information within an individual image, there are limitations on the capability of global feature discrimination.

\subsection{Contrastive learning for semantic segmentation}
Contrastive learning is a feature learning criterion that aims to minimize the distance between intra-class features while maximizing the distance between inter-class features. Recent advancements in contrastive learning with semantic segmentation~\cite{khosla2020supervised,wang2021exploring,pissas2022multi,hu2021region} have demonstrated impressive performances. 

Hu~\etal~\cite{hu2021region} and Wang~\etal~\cite{ wang2021exploring} proposed novel methods for semantic segmentation in a fully supervised setting that explores global pixel relations, extracting features from multiple images to regularize segmentation embedding space globally. Wang~\etal~\cite{wang2021exploring} introduced memory banks and segmentation-aware negative sampling methods, storing massive data to train distinctive representations and getting more gradient contributions during the training. Hu~\etal~\cite{hu2021region} introduced class-wise weighted region \rvc{centers that \rvvc{were generated} with positive samples, to \rvvc{be utilized} as the anchors in contrastive learning}. However, solely focusing on positive samples for weighting could diminish the discrimination ability of the model. Additionally, \cite{hu2021region, wang2021exploring} have neglected multiple scales of features except for the features of the last layer, \rvc{so they capture \rvvc{only} limited local/global contexts and relationships between local and global contexts.}

Finally, Pissas~\etal~\cite{pissas2022multi} proposed a method leveraging the multiple scales of features for supervised contrastive learning. That is, the researchers applied contrastive learning to the multi-scale and cross-scale features.
By doing so, \rvc{the method~\cite{pissas2022multi}} captures global/local context information from multi-scale features and the relationship \rvc{between local and global contexts} from cross-scale features. Nonetheless, it could not consistently grasp relationships across scales of features because multi-scale and cross-scale contrastive learning are operated separately. In some cases, features can be arranged differently in multi-scale and cross-scale contrastive learning. For example, a feature shifted by multi-scale contrastive learning can be shifted differently in cross-scale contrastive learning.

\section{\rvc{Contextrast: Contextual contrastive learning with BANE sampling}}
\label{sec:formatting}

\newcommand{\layerIdx}{\mathit{i}}
\newcommand{\layerTotal}{I}
\newcommand{\classIdx}{\mathit{n}}
\newcommand{\classTotal}{N}
\newcommand{\featureIdx}{\mathit{j}}

\newcommand{\featurevec}{\mathbf{f}}
\newcommand{\featureset}{\mathbf{F}}
\newcommand{\embeddingvec}{\mathbf{v}}
\newcommand{\embeddingset}{\mathbf{V}}

\newcommand{\anchor}{\mathbf{a}}
\newcommand{\anchors}{\mathbf{A}}

\newcommand{\gtfunc}{g}

\newcommand{\segGT}{\mathbf{Y}}
\newcommand{\segPred}{\hat{\segGT}}

\newcommand{\binaryImg}{\mathbf{B}}
\newcommand{\distImg}{\mathbf{D}}

\newcommand{\edgeSet}{\mathbf{E}}

\begin{figure*}[t]
    \centering
    \includegraphics[scale=0.50]{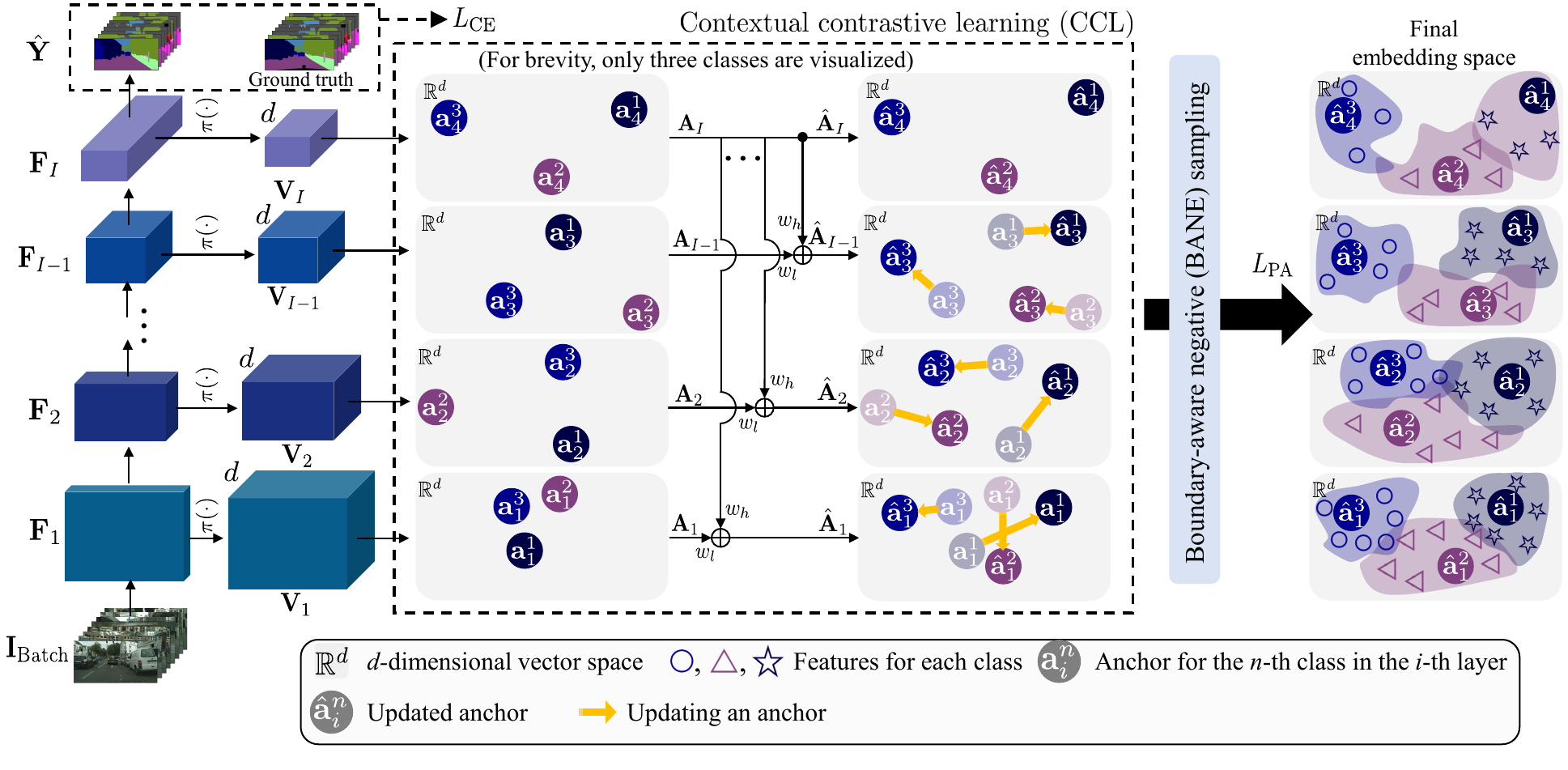}
    \caption{Overall Contextrast framework. Contextrast utilizes the representative anchors updated by the semantically rich representative anchor vector set $\anchors_\layerTotal$. Thus, it integrates local/global contexts and their relationships. Then, BANE sampling samples examples that exist along the boundaries of prediction error regions. It samples more informative negative samples and captures fine-grained details for contrastive learning.  $\mathbf{I}_\text{Batch}$ is the batch images. $\segPred$ is the prediction outcome from the model. $\featureset_\layerIdx$ is the feature map of the $\layerIdx$-th encoder layer. $\embeddingset_\layerIdx$ is the $\layerIdx$-th set of the embedded feature vector by the encoding function $\pi(\cdot)$. $\anchors_\layerIdx$ denotes the representative anchors of the $\layerIdx$-th embedded feature vector. The updated representative anchor $\hat{\anchors}_{\layerIdx}$ results from adding low-level and highest-level anchors. $w_h$ and $w_l$ are weight hyperparameters for updating representative anchors. The $L_{\text{PA}}$ is the proposed \abc{pixel-anchor} loss function. $L_{\text{CE}}$ represents the cross-entropy loss function. Features of each semantic class are \rvvc{illustrated in} different shapes and colors~(best viewed in color).}
    \label{fig:pipeline}
\end{figure*}

\subsection{Overall framework}
As shown in~\cref{fig:pipeline}, we propose a supervised contrastive learning framework encompassing two novel methods for semantic segmentation.


First, we propose a concept of \textit{representative anchors}, which are multi-scale-aware salient features implicitly representing the class by leveraging hierarchical design, as described in~\cref{sec:contextual_contrast}. Second, we deliberately sample the features corresponding to the boundaries within the regions that were incorrectly predicted as the negative samples, \abc{as described in~\cref{sec:boundary-aware}.}



\subsection{Contextual contrastive learning \rvvc{(CCL)}}
\label{sec:contextual_contrast}

Let us assume that an encoder consists of a total of $\layerTotal$ layers. 
Then, we begin by expressing the representative anchors corresponding to the $\layerIdx$-th encoder layer as $\anchors_\layerIdx$, where $\layerIdx \in \{1, \cdots, \layerTotal\}$. 
$\anchors_\layerIdx$ consists of $\classTotal$ class-wise representative anchors. 
Each anchor for the $\classIdx$-th class is denoted by $\anchor^\classIdx_\layerIdx \in \mathbb{R}^d$, which is defined as the average of embedded feature vectors belonging to the ground truth semantic class within the batch images as follows:

\begin{align}
    \anchor^\classIdx_\layerIdx = \frac{\sum\limits_{\embeddingvec \in \embeddingset_\layerIdx} \embeddingvec \mathbbm{1}[g(\embeddingvec) = n]}{\sum\limits_{\embeddingvec \in \embeddingset_\layerIdx} \mathbbm{1}[g(\embeddingvec) = n]}, \classIdx = 1, 2, ..., \classTotal,
    \label{eq:anchor}
\end{align}
where $\embeddingset_\layerIdx$ is an embedded feature vector set from the feature of the $\layerIdx$-th encoder layer's feature map $\featurevec \in \featureset_\layerIdx$, \abc{as illustrated in~\cref{fig:pipeline}}, i.e. $\embeddingvec = \pi(\featurevec)$; 
$\gtfunc(\cdot)$ represents a function that returns the ground truth semantic label of each embedding feature vector; 
$\mathbbm{1}[\cdot]$ is the Iverson bracket, which outputs one if the condition is satisfied and zero otherwise. 
By using \cref{eq:anchor}, $\anchors_\layerIdx$ is expressed as $\anchors_\layerIdx = \{\anchor^1_\layerIdx, \anchor^2_\layerIdx, ..., \anchor^\classTotal_\layerIdx\}$. For convenience, we interchangeably express $\anchors_\layerIdx$ in a matrix form, i.e. $\anchors_\layerIdx = [\anchor^1_\layerIdx \; \anchor^2_\layerIdx \; ... \;  \anchor^\classTotal_\layerIdx] \in \mathbb{R}^{d \times \classTotal}$.


Then, lower-level anchors $\anchors_\layerIdx$ are updated with the representative anchor of the last layer $\anchors_{\layerTotal}$ to encapsulate both high-level and low-level context, accounting for multi-scale.
Consequently, the updated representative anchor $\hat{\anchors}_\layerIdx$ is defined as $\hat{\anchors}_{\layerIdx}=w_l \anchors_\layerIdx + w_h \anchors_{\layerTotal}=\{\hat{\anchor}^1_\layerIdx, \hat{\anchor}^2_\layerIdx, ..., \hat{\anchor}^\classTotal_\layerIdx\}$, where $w_l$ and $w_h$ are weight hyperparameters for anchor update~(see ~\cref{fig:pipeline}). By updating the lower-level anchors, $\hat{\anchors}_\layerIdx$ can act as a criterion to capture relationships across different scales. For $i=\layerTotal$, $\hat{\anchors}_{\layerTotal}$ is defined as $\hat{\anchors}_{\layerTotal}={\anchors}_{\layerTotal}$. 
\begin{align}
      L_{\rvc{\mathrm{NCE}}}\!=\! \frac{-1}{|\embeddingset_{+}|}\!\sum\limits_{\embeddingvec_{+} \in \embeddingset_{+}}\!\log\!\frac{\exp(\embeddingvec\!\cdot\! \embeddingvec_+ / \tau)}{\exp(\embeddingvec\!\cdot\! \embeddingvec_+/ \tau) +\!\sum\limits_{\embeddingvec_{-}}\exp(\embeddingvec\!\cdot\! \embeddingvec_- / \tau)},
  \label{eq:info_nce}
\end{align}

Next, we incorporate \abc{InfoNCE~\cite{gutmann2010noise, oord2018representation} loss in~\cref{eq:info_nce}} with $\hat{\anchors}_{\layerIdx}$, which is referred to as the \textit{pixel-anchor~(PA) loss}, as follows:
\begin{align}
    L_{\mathrm{PA}} = \sum\limits_{\layerIdx=1}^{\layerTotal}\lambda_\layerIdx \biggl[\frac{1}{\classTotal}\sum_{\hat{\anchor}^\classIdx_\layerIdx\in\hat{\anchors}_\layerIdx}\frac{-1}{|\embeddingset_+|}\sum_{\embeddingvec_+\in\embeddingset_+} L_a\biggr]
    \label{eq:pixel-anchor}
\end{align}
\begin{align}
    L_a = \log\frac{\exp(\hat{\anchor}^\classIdx_\layerIdx\!\cdot\!\embeddingvec_+/\tau)}{\exp(\hat{\anchor}^\classIdx_\layerIdx\!\cdot\!\embeddingvec_+/\tau)\!+\! \!\sum\limits_{\embeddingvec_-\in\embeddingset_-}\exp(\hat{\anchor}^\classIdx_\layerIdx\!\cdot\!\embeddingvec_-/\tau)}
    \label{eq:pa-for-each-i}
\end{align}
where \abc{$\embeddingvec_{+/-}$ represents positive and negative samples, respectively, and} $\lambda_\layerIdx$ represents the weight hyperparameters assigned to the pixel-anchor contrastive loss for the $\layerIdx$-th encoder layer.
While the anchor is set with individual features in \cref{eq:info_nce}, the anchor is set with a representative anchor $\hat{\anchors}_\layerIdx$ in \abc{\cref{eq:pa-for-each-i}}. The pixel-anchor loss aims to optimize embedding features by minimizing the distance between intra-class features and their corresponding representative anchors while maximizing the separation between inter-class features and their corresponding representative anchors. Thus, the network captures global context and intricate details from multi-scale features and their connection using the representative anchors as the criterion.

Furthermore, the pixel-anchor loss operates in conjunction with the conventional pixel-wise cross-entropy loss~$L_{\text{CE}}$~\cite{sun2019high}, providing a complementary approach to enhance segmentation performance. 
This synergy is purposeful: while pixel-wise cross-entropy loss aims to predict the correct label for each sample, pixel-anchor loss aims to learn good data representations by considering the relationships between different samples.


As a result, the primary objective of the entire framework is to optimize the following loss:
\begin{equation}
    L = L_{\mathrm{CE}} +  \alpha L_{\mathrm{PA}},
\end{equation}
where $\alpha$ represents the weight for our pixel-anchor loss.

\begin{figure}[t]
    \centering
    \begin{subfigure}[h]{0.5\textwidth}
        \centering
        \includegraphics[scale=0.43]{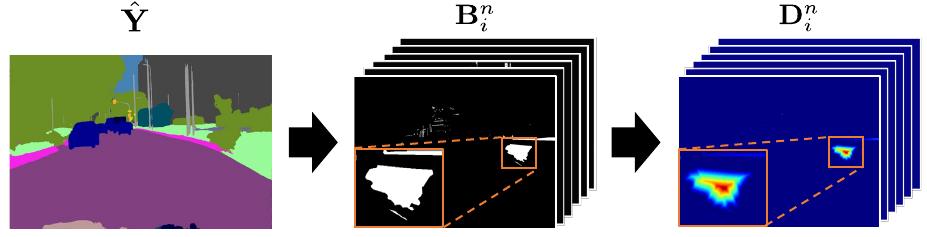}
        \caption{}
        \label{fig:PP}
    \end{subfigure}
    \begin{subfigure}[h]{0.5\textwidth}
        \centering
        \includegraphics[scale=0.27]{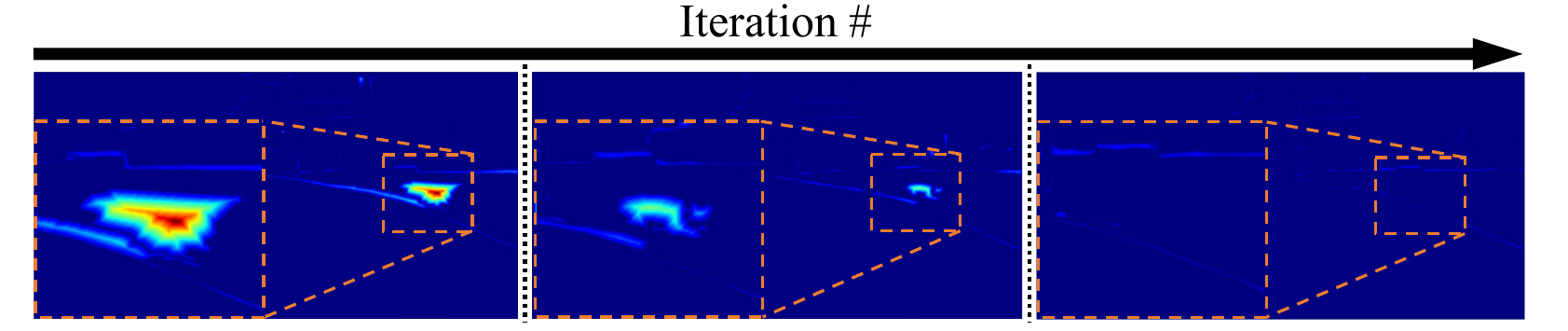}
        \caption{}
        \label{fig:PA}
    \end{subfigure}
    \caption{ Visual description of boundary-aware negative sampling and how the under/over-segmentation problems are addressed during the training. (a) The prediction outcome $\segPred$ is decomposed into class-wise binary maps $\textbf{B}^\classIdx_\layerIdx$. Then, class-wise distance maps $\textbf{D}^\classIdx_\layerIdx$ are generated with the Distance Transform~\cite{kimmel1996sub}. (b)~The evolution of the distance map over iterations. The wrongly predicted regions shrink during training~(best viewed in color).}
    \label{fig:ext_sampling}
\end{figure}

\subsection{Boundary-aware negative~(BANE) sampling}
\label{sec:boundary-aware}
While enhancing the loss function, we also propose an effective negative sampling approach that considers the boundaries of the prediction error to increase the quality of $\embeddingvec_{-}$ in \abc{\cref{eq:pa-for-each-i}}. \abc{To do that, we incorporate a simple but effective boundary extraction method from $\mathrm{\hat{Y}}$~\cite{yuan2020segfix,wang2022active}.}
The method mainly consists of three steps as illustrated in \cref{fig:ext_sampling}:~1)~decomposing prediction output to class-wise binary error maps, 2) distance transform based on the class-wise error maps, and 3) selecting negative samples.

To extract negative samples, the class-wise binary error map $\binaryImg^\classIdx_\layerIdx$ for each pixel $(u, v)$ is defined as: 
\begin{equation}
\binaryImg^\classIdx_\layerIdx(u,v)=\mathbbm{1}[(\hat{{y}_i} \neq \classIdx) \land (\gtfunc(\hat{{y}}_i) = \classIdx)],
\end{equation}
where $\hat{y}_i$ denotes the predicted class for the $i$-th layer downsampled from the predicted class in the final layer. $\gtfunc$($\cdot$) denotes a labeling function in the same way as in~\cref{sec:contextual_contrast}. The $\binaryImg^\classIdx_\layerIdx$ has a value of one for the wrongly predicted pixel, i.e.~a negative sample, and zero otherwise, as shown in~\cref{fig:ext_sampling}(a).

Next, $\binaryImg^\classIdx_\layerIdx$ is converted to a class-wise distance map $\distImg^\classIdx_\layerIdx$ by the distance transform~\cite{kimmel1996sub}. The pixel value of $\distImg^\classIdx_\layerIdx$ is the minimal distance between the pixel $(u,v)$ and the edge pixels~$\edgeSet^\classIdx_\layerIdx$, which is from the corresponding class-wise error map and is defined as follows:
\begin{equation}
    \distImg^\classIdx_\layerIdx(u,v) = \mathop{\mathrm{min}}_{(x,y)\in \textbf{E}^\classIdx_\layerIdx}\sqrt{(u-x)^2+(v-y)^2}.
\end{equation}
This implies that within the incorrectly predicted regions, where the pixel value of $\binaryImg^\classIdx_\layerIdx$ is one~(white regions in \cref{fig:ext_sampling}(a)), a lower value indicates a higher probability of the pixel being on the boundary.

Finally, among the regions whose values in $\binaryImg^\classIdx_\layerIdx$ are one, we select embedding vectors corresponding to the lower $K$ percentage of the smallest distances in $\distImg^\classIdx_\layerIdx$ as negative samples for each $n$-th representative anchor in \cref{eq:pa-for-each-i}.
\rvc{By incorporating these vectors into~\cref{eq:pixel-anchor}, \rvvc{Contextrast} allows these vectors to be close to the anchor of the true class and far from the features of incorrectly predicted classes during training. Thus, these boundary-aware negative samples help the network learn the inter-spatial relationship between the segmentation classes \rvvc{better}.}

\section{Experiments}
\subsection{Experimental setup}\label{sec:exp-setup}
{\bf Datasets.} We conduct our experiments using five public datasets: Cityscapes~\cite{cordts2016cityscapes}, ADE20K~\cite{zhou2017scene}, PASCAL-C~\cite{mottaghi2014role}, COCO-Stuff~\cite{caesar2018coco}, and CamVid~\cite{brostow2009semantic} datasets. For a fair comparison, we follow the existing training and validation settings of the datasets~(details are explained in the supplementary materials). 
Among them, because the Cityscapes dataset additionally provides the public benchmark by using test data, we differentiate \rvc{the} validation and test sets using the suffix \texttt{test}, \rvc{i.e.}~Cityscapes-\texttt{test}.


\noindent{\bf Training settings.} To demonstrate the \rvc{efficacy} of our proposed approach, we employ three networks: a)~DeepLabV3~\cite{chen2017rethinking},~b) HRNet~\cite{sun2019high}, and c)~OCRNet~\cite{yuan2020object}. D-ResNet-101 backbone is utilized in DeepLabV3. HRNetV2-W48 backbone is employed in HRNet and OCRNet networks.
 We \rvc{used same} hyperparameters and initialized \rvc{the network} using pre-trained \rvc{weights} on ImageNet~\cite{deng2009imagenet} while the remaining layers were randomly initialized. We utilized color jittering, horizontal flipping, and random scaling for data augmentation. Stochastic gradient descent~(SGD) is applied as an optimizer for CNN backbones with a momentum of 0.9. 
 In addition, polynomial annealing policy~\cite{chen2017rethinking} is applied to schedule the learning rate, which is multiplied by ${(1-\frac{\text{Iteration \#}}{\text{Total\ iterations}})^{0.9}}$. 
 \abc{$\lambda_{4\rightarrow1}$ is set to 1.0, 0.7, 0.4, and 0.1. $\alpha$ is set to 0.1.}
  On the Cityscapes dataset, we have set a batch size of 8 for 40K iterations and \rvc{cropped} from 1024$\times$2048 to 512$\times$1024. The model is trained on the CamVid dataset with a batch size of 16 for 6K iterations. On ADE20K, the models are trained with a crop size of 512$\times$512 and a batch size of 12 for 80K iterations. On COCO-Stuff and PASCAL-C, the models are trained with a crop size of 512$\times$512 and \rvc{a} batch size of 16 for 60K iterations. Note that we do not use any extra training data.


\noindent \textbf{Testing settings.}  We follow the general setup~\cite{sun2019high,yuan2020object,chen2017rethinking}, averaging the segmentation results over multiple scales with flipping for CamVid, COCO-Stuff, ADE20K, \rvc{and} PASCAL-C datasets. The scaling factor is set from 0.75 to 2.0 with intervals of 0.25. We employed single-scale evaluation for Cityscapes to follow the experimental setup of \rvvc{multi/cross-scale contrastive learning}~\cite{pissas2022multi}.

\begin{figure*}[t!]
    \centering
    \includegraphics[scale=0.45]{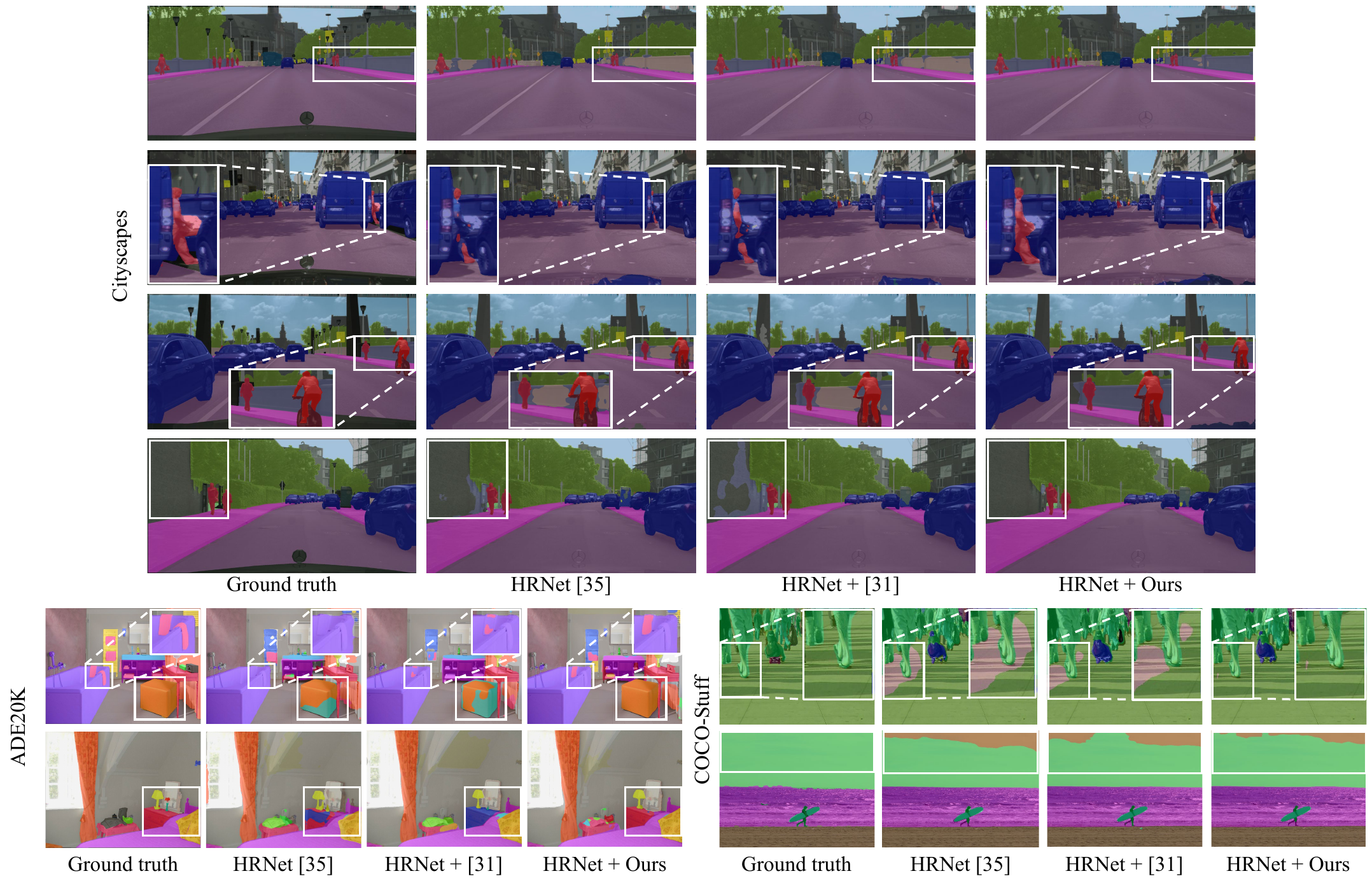}
    \caption{Qualitative results \rvc{from} HRNet~\cite{sun2019high}, HRNet +~\cite{pissas2022multi}, and HRNet + Ours on the Cityscapes, ADE20K, and COCO-Stuff datasets, respectively~(best viewed on color).}
    \label{fig:qual_city}
\end{figure*}

\begin{table*}[t!]
\scriptsize
\centering
\renewcommand{\arraystretch}{1.3}
\begin{tabular}{llc|ccccc} 
\hline
\rowcolor[HTML]{DAE8FC} 
\multicolumn{1}{c|}{\cellcolor[HTML]{DAE8FC}}  &                             \multicolumn{2}{c|}{\cellcolor[HTML]{DAE8FC}Description} & \multicolumn{5}{c}{\cellcolor[HTML]{DAE8FC}Dataset [mIoU (\%)]}  \\  \cline{2-3} \cline{4-8} 
\rowcolor[HTML]{DAE8FC} 
\multicolumn{1}{c|}{\multirow{-2}{*}{\cellcolor[HTML]{DAE8FC}Method}} & \multicolumn{1}{c}{Loss} & Sampling & Cityscapes                                       & CamVid  & COCO-Stuff & ADE20K & PASCAL-C \\ \hline
\rowcolor[HTML]{FFFFFF} 
\multicolumn{1}{l|}{\cellcolor[HTML]{FFFFFF}DeepLabV3} & \multicolumn{1}{l}{\cellcolor[HTML]{FFFFFF}$L_{\rvc{\mathrm{CE}}}$}      & None & {\cellcolor[HTML]{FFFFFF}77.12} & {\cellcolor[HTML]{FFFFFF}78.80} & {\cellcolor[HTML]{FFFFFF}37.92}          & {42.85} & {52.01} \\  
\multicolumn{1}{l|}{\cellcolor[HTML]{FFFFFF}DeepLabV3 + \cite{pissas2022multi}} & \multicolumn{1}{l}{\cellcolor[HTML]{FFFFFF}$L_{\rvc{\mathrm{CE}}} + L_{\rvc{\mathrm{cms}}} + L_{\rvc{\mathrm{ccs}}}$}      & Random & {\cellcolor[HTML]{FFFFFF}78.94} \color[HTML]{2D8C00}{(+1.82)} & {79.67} {\color[HTML]{2D8C00} {(+0.87)}} & {37.39} \color[HTML]{8A0101}{(-0.53)}          & 43.86 \color[HTML]{2D8C00}{(+1.01)} & {51.52} \color[HTML]{8A0101}{(-0.49)} \\
\rowcolor[HTML]{EFEFEF}
\multicolumn{1}{l|}{DeepLabV3 + Ours} & \multicolumn{1}{l}{$L_{\rvc{\mathrm{CE}}} + L_{\rvc{\mathrm{PA}}}$ (Ours)}      & Boundary-aware (Ours) & {\textbf{79.35}} \color[HTML]{2D8C00}{\textbf{(+2.23)}} & \textbf{79.98} {\color[HTML]{2D8C00} {\textbf{(+1.18)}}} & \textbf{38.12} \color[HTML]{2D8C00}{\textbf{(+0.20)}}          & \textbf{44.12} \color[HTML]{2D8C00}{\textbf{(+1.27)}} & \textbf{52.62} \color[HTML]{2D8C00}{\textbf{(+0.61)}} \\ \hline
\rowcolor[HTML]{FFFFFF} 
\multicolumn{1}{l|}{\cellcolor[HTML]{FFFFFF}HRNet} & \multicolumn{1}{l}{\cellcolor[HTML]{FFFFFF}$L_{\rvc{\mathrm{CE}}}$}      & None & {\cellcolor[HTML]{FFFFFF}78.48} & {\cellcolor[HTML]{FFFFFF}82.17} & {\cellcolor[HTML]{FFFFFF}36.04}          & {41.86} & {51.86} \\  
\multicolumn{1}{l|}{\cellcolor[HTML]{FFFFFF}HRNet + \cite{wang2021exploring}} & \multicolumn{1}{l}{\cellcolor[HTML]{FFFFFF}$L_{\rvc{\mathrm{CE}}} + L_{\rvc{\mathrm{NCE}}}$}      & Semi-hard & {\cellcolor[HTML]{FFFFFF}81.00} \color[HTML]{2D8C00}{(+2.52)} & N/A & N/A  & N/A & N/A \\
\multicolumn{1}{l|}{\cellcolor[HTML]{FFFFFF}HRNet + \cite{hu2021region}} & \multicolumn{1}{l}{\cellcolor[HTML]{FFFFFF}$L_{\rvc{\mathrm{CE}}} + L_{\rvc{\mathrm{NCE}}} + L_{\rvc{\mathrm{Aux}}}$}      & Random & {\cellcolor[HTML]{FFFFFF}81.90} \color[HTML]{2D8C00}{(+3.42)} & N/A & N/A  & N/A & N/A \\
\multicolumn{1}{l|}{\cellcolor[HTML]{FFFFFF}HRNet + \cite{pissas2022multi}} & \multicolumn{1}{l}{\cellcolor[HTML]{FFFFFF}$L_{\rvc{\mathrm{CE}}} + L_{\rvc{\mathrm{cms}}} + L_{\rvc{\mathrm{ccs}}}$}      & Random & {\cellcolor[HTML]{FFFFFF}81.50} \color[HTML]{2D8C00}{(+3.02)} & {83.14} {\color[HTML]{2D8C00} {(+0.97)}} & \textbf{36.35} \color[HTML]{2D8C00}{\textbf{(+0.31)}}          & {43.27} \color[HTML]{2D8C00}{(+1.41)} & {52.11} \color[HTML]{2D8C00}{(+0.25)} \\
\rowcolor[HTML]{EFEFEF}
\multicolumn{1}{l|}{HRNet + Ours} & \multicolumn{1}{l}{$L_{\rvc{\mathrm{CE}}} + L_{\rvc{\mathrm{PA}}}$ (Ours)}      & Boundary-aware (Ours) & {\textbf{82.20}} \color[HTML]{2D8C00}{\textbf{(+3.72)}} & \textbf{84.33} {\color[HTML]{2D8C00} {\textbf{(+2.16)}}} & 36.34 \color[HTML]{2D8C00}{(+0.30)}          & \textbf{43.42} \color[HTML]{2D8C00}{\textbf{(+1.56)}} & \textbf{52.17} \color[HTML]{2D8C00}{\textbf{(+0.31)}} \\ \hline
\rowcolor[HTML]{FFFFFF} 
\multicolumn{1}{l|}{\cellcolor[HTML]{FFFFFF}OCRNet} & \multicolumn{1}{l}{\cellcolor[HTML]{FFFFFF}$L_{\rvc{\mathrm{CE}}}$}      & None & {\cellcolor[HTML]{FFFFFF}79.95} & {\cellcolor[HTML]{FFFFFF}82.69} & {\cellcolor[HTML]{FFFFFF}39.00}          & {41.51} & {54.35} \\  
\multicolumn{1}{l|}{\cellcolor[HTML]{FFFFFF}OCRNet + \cite{pissas2022multi}} & \multicolumn{1}{l}{\cellcolor[HTML]{FFFFFF}$L_{\rvc{\mathrm{CE}}} + L_{\rvc{\mathrm{cms}}} + L_{\rvc{\mathrm{ccs}}}$}      & Random & {\cellcolor[HTML]{FFFFFF}81.51} \color[HTML]{2D8C00}{(+1.56)} & {83.82} {\color[HTML]{2D8C00} {(+1.13)}} & {38.55} \color[HTML]{8A0101}{(-0.45)}          & {43.28} \color[HTML]{2D8C00}{(+1.77)} & {54.48} \color[HTML]{2D8C00}{(+0.13)}\\
\rowcolor[HTML]{EFEFEF}
\multicolumn{1}{l|}{\cellcolor[HTML]{EFEFEF}OCRNet + Ours} & \multicolumn{1}{l}{$L_{\rvc{\mathrm{CE}}} + L_{\rvc{\mathrm{PA}}}$ (Ours)}      & Boundary-aware (Ours) & {\textbf{81.94}} \color[HTML]{2D8C00}{\textbf{(+1.99)}} & \textbf{84.10} {\color[HTML]{2D8C00} {\textbf{(+1.41)}}} & \textbf{39.08} \color[HTML]{2D8C00}{\textbf{(+0.08)}}          & \textbf{43.84} \color[HTML]{2D8C00}{\textbf{(+2.33)}} & \textbf{54.64} \color[HTML]{2D8C00}{\textbf{(+0.29)}} \\ \hline
\end{tabular}

\caption{Quantitative results on public datasets compared with the state-of-the-art contrastive learning-based semantic segmentation approaches. We employed DeepLabV3~\cite{chen2017rethinking}, HRNet~\cite{sun2019high}, and OCRNet~\cite{yuan2020object} as segmentation models.} 
\label{tab:comparison_all}
\end{table*}

\noindent {\bf Evaluation metric.} In the experiments, we quantitatively analyze the performance with respect to a)~semantic segmentation results and b)~the distinctiveness of the features. 

To evaluate the semantic segmentation performance, the mean of class-wise intersection over union (mIoU)~\cite{wang2021exploring,hu2021region} is used as an evaluation metric. For the Cityscapes-\texttt{test}, an instance-level intersection-over-union metric~(iIoU)~\cite{cordts2016cityscapes} is also used to evaluate how the individual instances are well-segmented.
This is because mIoU can be biased toward object instances that cover a large image area in the street scenes. 
The iIoU is defined as \rvc{follows}:
\begin{equation}
    \text{iIoU} = \frac{\text{iTP}}{(\text{iTP} + \text{FP} + \text{iFN})},
\end{equation}
where iTP, FP, and iFN denote the numbers of true positive, false positive, and the number of false negative pixels, respectively. Note that iTP and iFN are calculated with \rvc{weighted} pixel contributions based on the ratio of each class's average instance size to the corresponding ground truth instance size. 

Next, for the feature-level analysis, we adopt the following three metrics: intra-class alignment~(A) to evaluate how well the intra-class features are closely clustered, 
inter-class uniformity~(U) to evaluate how far the centroids of features originating from different classes are separated in the embedding space,
and inter-class neighborhood uniformity~(U$_l$) to measure the separation of the $l$-closest centroids of inter-class features, which indicates how clearly the decision boundaries are \rvc{discriminated} between the $l$-closest centroids. More details can be found in~\cite{li2022targeted} (see~\cref{sec:feature_level}).


\subsection{Semantic segmentation performance}

The first experiment \rvc{compares} the performance of our proposed approach \rvc{with that of the} existing contrastive learning-based methods, \rvc{to support} the claim
that our approach enables networks to output more precise semantic segmentation, particularly 
resolving under- and over-segmentation issues. 
For our comparison, we used the following existing contrastive learning-based approaches: \rvvc{ContrastiveSeg}~\cite{wang2021exploring} that incorporates $L_{\mathrm{CE}}$ with $L_{\mathrm{NCE}}$\rvc{;} semi-hard negative sampling that just randomly selects the negative samples corresponding to the wrongly predicted regions\rvc{;} \rvc{region-aware contrastive learning}~\cite{hu2021region} \rvc{that} additionally employs auxiliary loss \rvc{$L_{\text{Aux}}$;} \rvc{multi/cross-scale contrastive learning}~\cite{pissas2022multi} that exploits multi-scale and cross-scale contrastive loss terms, \rvc{i.e.} $L_{\mathrm{cms}}$ and $L_{\mathrm{ccs}}$.

As shown in Fig.~\ref{fig:qual_city}, \rvc{~\Cref{tab:comparison_all}}, and \rvc{\Cref{tab:city_test}},
the state-of-the-art methods showed precise semantic segmentation results, mostly improving the mIoU compared with the baseline segmentation networks. 
Our proposed method exhibits noticeable improvements \rvc{on} the public datasets, mostly achieving the highest mIoU.
In particular, our proposed method even resolved the under- and over-segmentation more clearly~(see Fig.~\ref{fig:qual_city}).

In addition to the substantial performance improvement, we specifically focus on the differences in the loss terms.
As presented in \rvc{\Cref{tab:comparison_all}}, the performance after the application \rvc{of} the auxiliary loss~$L_{\rvc{\mathrm{Aux}}}$~\cite{hu2021region} showed higher mIoU compared with \rvvc{ContrastiveSeg}~\cite{wang2021exploring}, which only \rvc{incorporates} $L_{\rvc{\mathrm{CE}}}$ with $L_{\rvc{\mathrm{NCE}}}$. The multi-scale and cross-scale contrastive losses~\cite{pissas2022multi}, which are improved versions of $L_{\rvc{\mathrm{NCE}}}$ in a different scale level, also showed a substantial increase in mIoU. However, our method showed a large performance increase. 

In particular, it is noticeable that both \rvc{multi/cross-scale contrastive learning}~\cite{pissas2022multi} and ours considered multi-scale, yet \rvc{our} approach showed more stable performance increase. Occasionally, \rvc{multi/cross-scale contrastive learning}~\cite{pissas2022multi} \rvc{showed the degraded} performance owing to the conflict of the influences of two disentangled loss terms. 
That is, considering multi-scale and cross-scale separately in the learning process can sometimes result in the direction in which the \rvc{moved} embedding vector \rvc{becomes} undesirable, leading to a situation where the distinctiveness of vectors in the embedded space may not significantly increase.
\abc{Furthermore, the proposed method significantly improved boundary mIoU (B-mIoU) by incorporating BANE sampling into the contrastive learning, as demonstrated in~\cref{tab:b-miou}.}


Therefore, we conclude that our Contextrast is more effective for accurate semantic segmentation than \rvc{the} existing methods.


\begin{table}[]
\scriptsize
\centering
\renewcommand{\arraystretch}{1.3}
\begin{tabular}{l|cccc}
\hline
\rowcolor[HTML]{DAE8FC} 
\multicolumn{1}{c|}{\cellcolor[HTML]{DAE8FC}}                         & \multicolumn{2}{c}{\cellcolor[HTML]{DAE8FC}Classes}                               & \multicolumn{2}{c}{\cellcolor[HTML]{DAE8FC}Categories}                            \\ \cline{2-5} 
\rowcolor[HTML]{DAE8FC} 
\multicolumn{1}{c|}{\multirow{-2}{*}{\cellcolor[HTML]{DAE8FC}Method}} & mIOU (\%)                               & iIOU (\%)                               & mIOU (\%)                               & iIOU (\%)                                \\ \hline
\\[-0.5ex]
\multirow{-2}{*}{HRNet}                                               & \multirow{-2}{*}{79.51}                 & \multirow{-2}{*}{57.96}                 & \multirow{-2}{*}{91.33}                 & \multirow{-2}{*}{80.29}                 \\[-1.0ex]
                                                                      & 80.12                                   & 59.04                                   & 91.42                                   & 81.64                                   \\[-0.7ex]
\multirow{-2}{*}{HRNet + \cite{pissas2022multi}}                                               & {\color[HTML]{2D8C00} (+0.61)}          & {\color[HTML]{2D8C00} (+1.08)}          & {\color[HTML]{2D8C00} (+0.09)}          & {\color[HTML]{2D8C00} (+1.35)}          \\
\rowcolor[HTML]{EFEFEF} 
\cellcolor[HTML]{EFEFEF}                                              & \textbf{80.39}                          & \textbf{61.06}                                   & \textbf{91.59}                          & \textbf{82.14}                          \\[-0.7ex]
\rowcolor[HTML]{EFEFEF} 
\multirow{-2}{*}{\cellcolor[HTML]{EFEFEF}HRNet + Ours}                & {\color[HTML]{2D8C00} \textbf{(+0.88)}} & {\color[HTML]{2D8C00} \textbf{(+3.10)}}          & {\color[HTML]{2D8C00} \textbf{(+0.26)}} & {\color[HTML]{2D8C00} \textbf{(+1.85)}} \\ \hline
                                                                      &                                         &                                         &                                         &                                         \\[-0.5ex]
\multirow{-2}{*}{OCRNet}                                              & \multirow{-2}{*}{80.64}                 & \multirow{-2}{*}{58.72}                 & \multirow{-2}{*}{91.41}        & \multirow{-2}{*}{80.77}                 \\[-1.0ex]
\rowcolor[HTML]{EFEFEF} 
\cellcolor[HTML]{EFEFEF}                                              & \textbf{81.94}                          & \textbf{62.11}                          & \textbf{91.60}                                   & \textbf{81.83}                          \\[-0.7ex]
\rowcolor[HTML]{EFEFEF} 
\multirow{-2}{*}{\cellcolor[HTML]{EFEFEF}OCRNet + Ours}               & {\color[HTML]{2D8C00} \textbf{(+1.30)}} & {\color[HTML]{2D8C00} \textbf{(+3.39)}} & {\color[HTML]{2D8C00} \textbf{(+0.19)}}          & {\color[HTML]{2D8C00} \textbf{(+1.06)}} \\ \hline
                                                                      &                                         &                                         &                                         &                                         \\[-0.5ex]
\multirow{-2}{*}{DeeplabV3}                                          & \multirow{-2}{*}{77.01}                 & \multirow{-2}{*}{55.56}                 & \multirow{-2}{*}{89.64}                 & \multirow{-2}{*}{77.42}                 \\[-1.0ex]
\rowcolor[HTML]{EFEFEF} 
\cellcolor[HTML]{EFEFEF}                                              & \textbf{78.23}                          & \textbf{56.83}                          & \textbf{89.86}                          & \textbf{77.93}                          \\[-0.7ex]
\rowcolor[HTML]{EFEFEF} 
\multirow{-2}{*}{\cellcolor[HTML]{EFEFEF}DeeplabV3 + Ours}           & {\color[HTML]{2D8C00} \textbf{(+1.22)}} & {\color[HTML]{2D8C00} \textbf{(1.27)}}  & {\color[HTML]{2D8C00} \textbf{(+0.22)}} & {\color[HTML]{2D8C00} \textbf{(+0.51)}} \\ \hline
\end{tabular}
\caption{Quantitative segmentation results on Cityscapes-\texttt{test}.}
\label{tab:city_test}
\end{table}

\begin{table}[t]
\scriptsize
\centering
\renewcommand{\arraystretch}{1.3}
\begin{tabular}{c|ccc}
\hline
\rowcolor[HTML]{DAE8FC} 
        & B-mIoU (5px) & B-mIoU (7px) & B-mIoU (10px) \\ \hline
HRNet   & 59.93        & 65.82             & 69.25              \\
HRNet +~\cite{pissas2022multi} & 60.44 \color[HTML]{2D8C00}(+0.51)        &  66.29 (\color[HTML]{2D8C00}+0.47)            & 69.65 (\color[HTML]{2D8C00}+0.4)             \\
\rowcolor[HTML]{EFEFEF}
Ours    & \textbf{61.76 (\color[HTML]{2D8C00}+1.83)}        & \textbf{67.58 (\color[HTML]{2D8C00}+1.76)}            & \textbf{70.93 (\color[HTML]{2D8C00}+1.68)}              \\ \hline
\end{tabular}
\caption{\abc{B-mIoU performance comparison with Cityscapes dataset. B-mIoU represents the mIoU of the boundary region which is within pixels from the boundary.}}
\label{tab:b-miou}
\end{table}

\subsection{Feature-level in-depth analyses}\label{sec:feature_level}

Furthermore, we conducted two experiments to demonstrate that our Contextrast enhances the distinctiveness of the vectors on the embedded feature space. 

First, we assessed how the embedding vectors are aligned in the last layer just before reaching the segmentation head by using feature-level metrics, which \rvc{were} explained in \cref{sec:exp-setup}.
As presented in \rvc{\Cref{tab:alginment}}, we demonstrate that our proposed method aligns intra-class features and pushes away inter-class features, improving all the metrics. 
Thus, it implies that our method makes the model have distinctive decision boundaries because intra-class features are well-organized and inter-class features are well-discriminated in the latent space.

\begin{figure}[h]
    \centering
    \begin{subfigure}[h]{0.20\textwidth}
        \centering
        \includegraphics[scale=0.165]{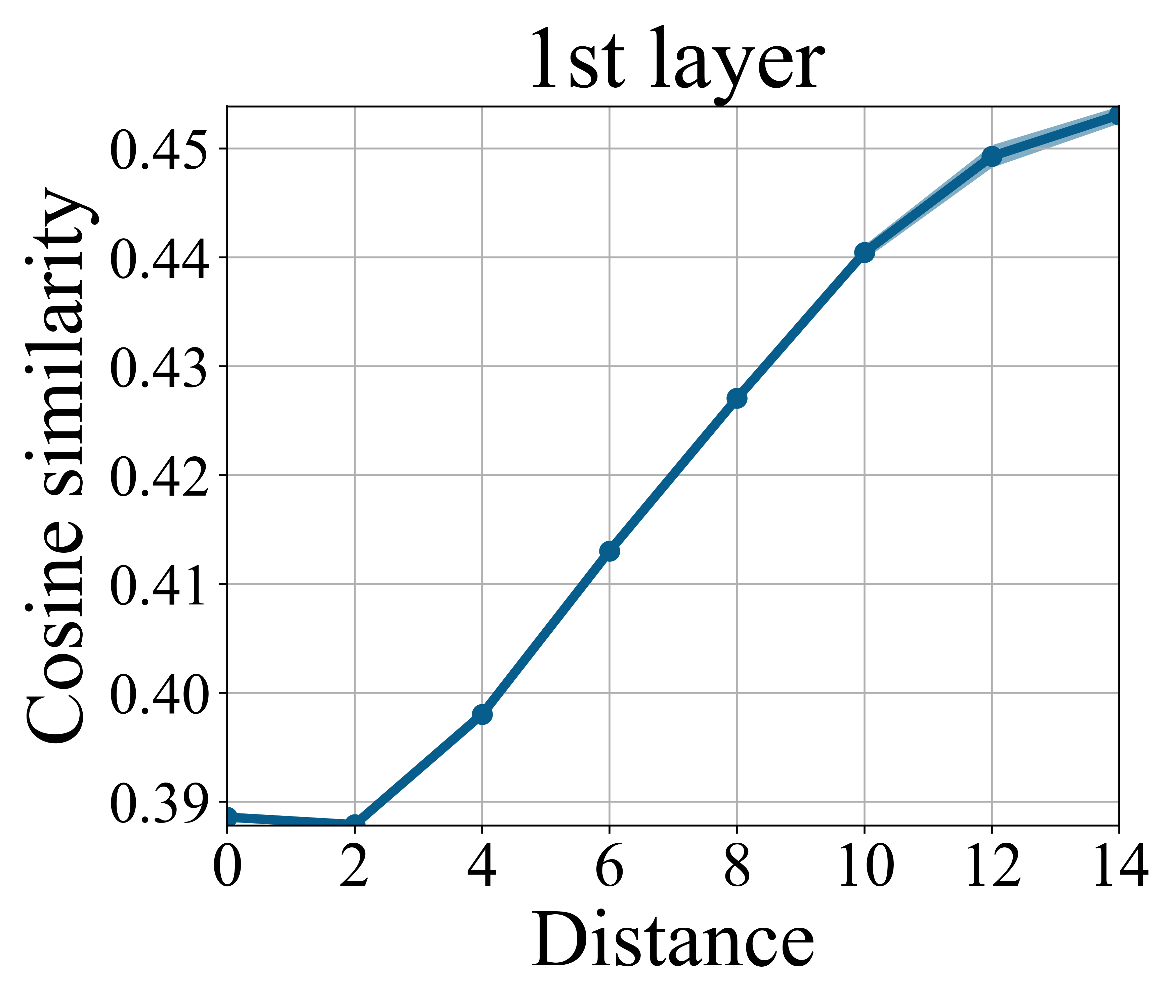}
        \label{fig:scale0_cossim}
    \end{subfigure}
    \begin{subfigure}[h]{0.20\textwidth}
        \centering
        \includegraphics[scale=0.165]{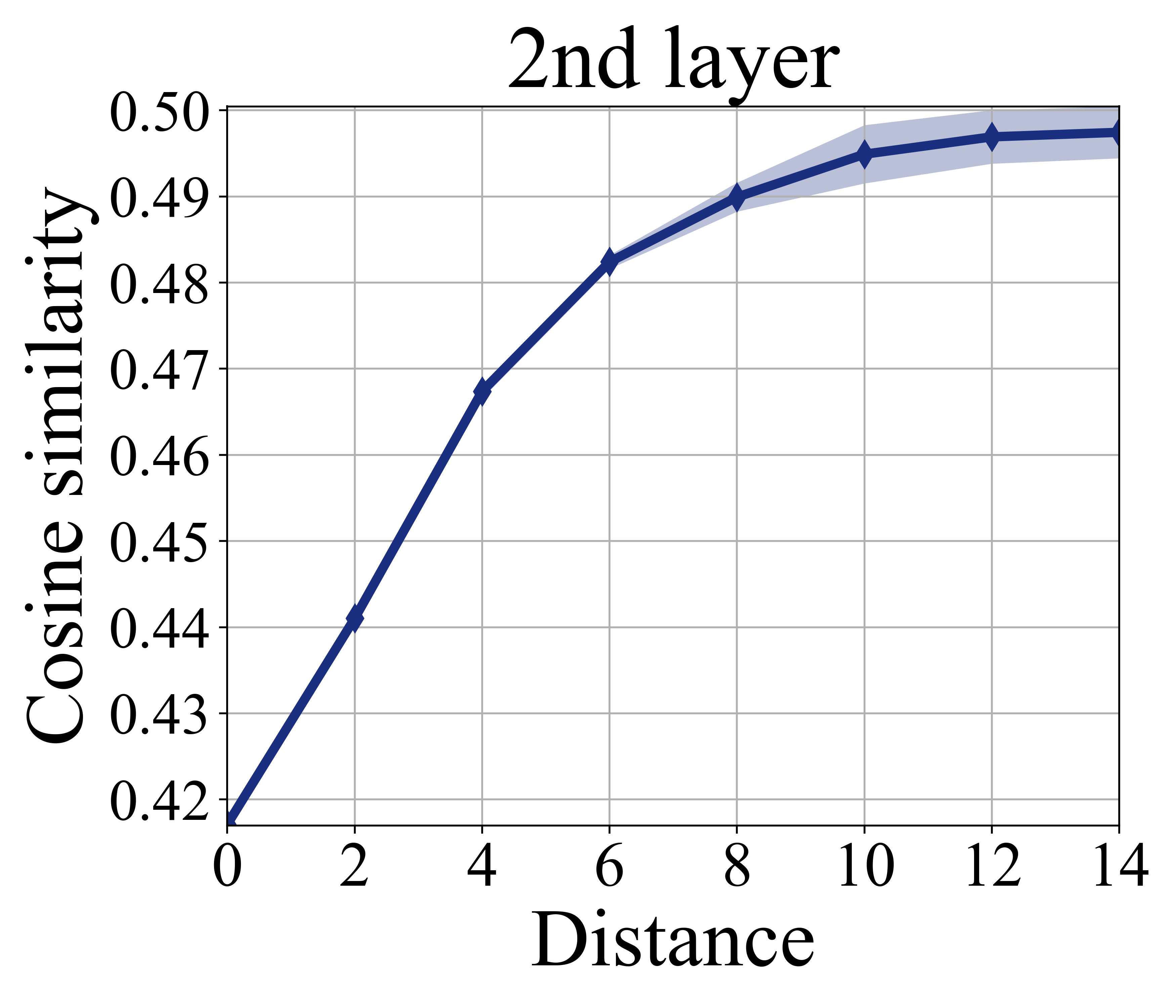}
        \label{fig:scale1_cossim}
    \end{subfigure}
    \begin{subfigure}[h]{0.20\textwidth}
        \centering
        \includegraphics[scale=0.165]{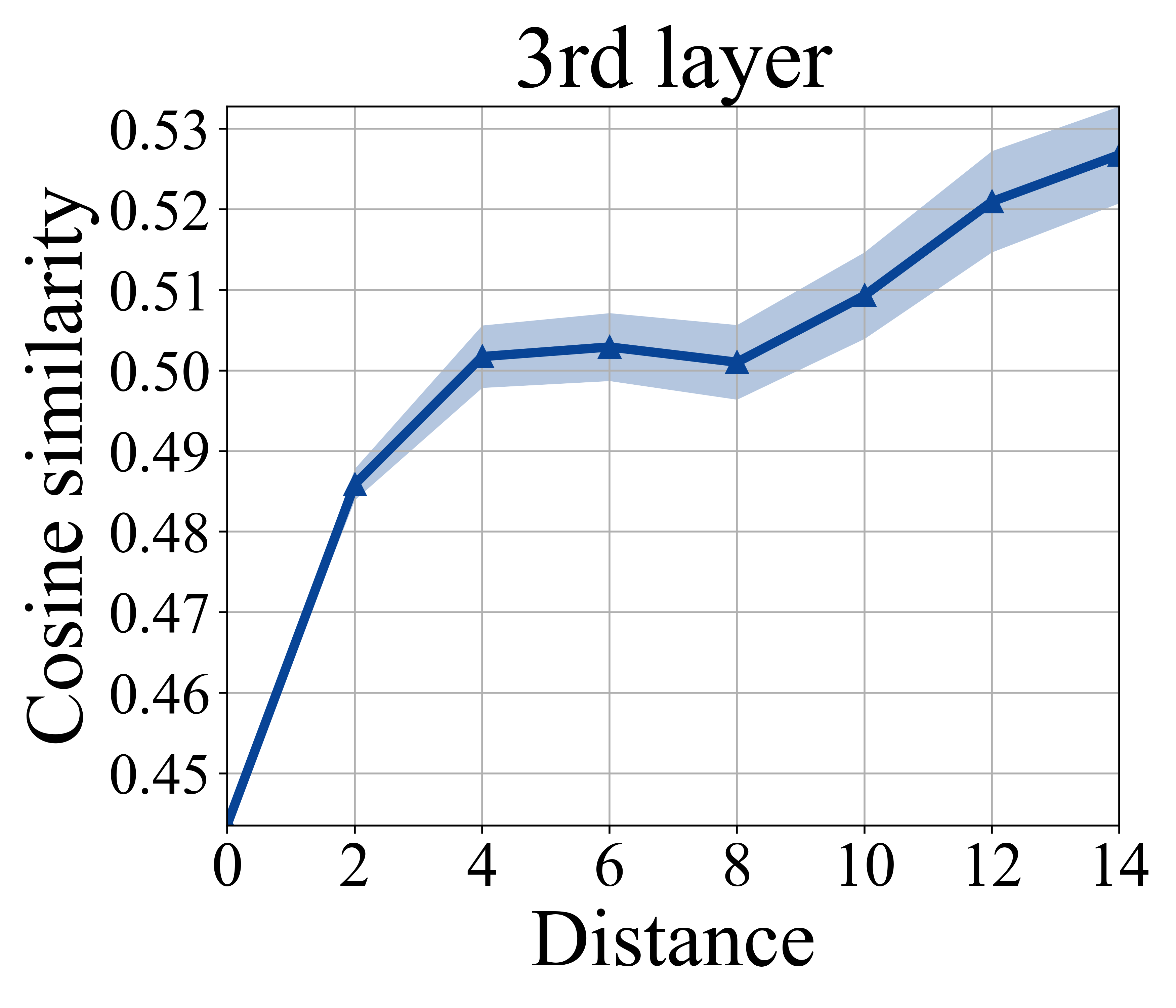}
        \label{fig:scale2_cossim}
    \end{subfigure}
    \begin{subfigure}[h]{0.20\textwidth}
        \centering
        \includegraphics[scale=0.165]{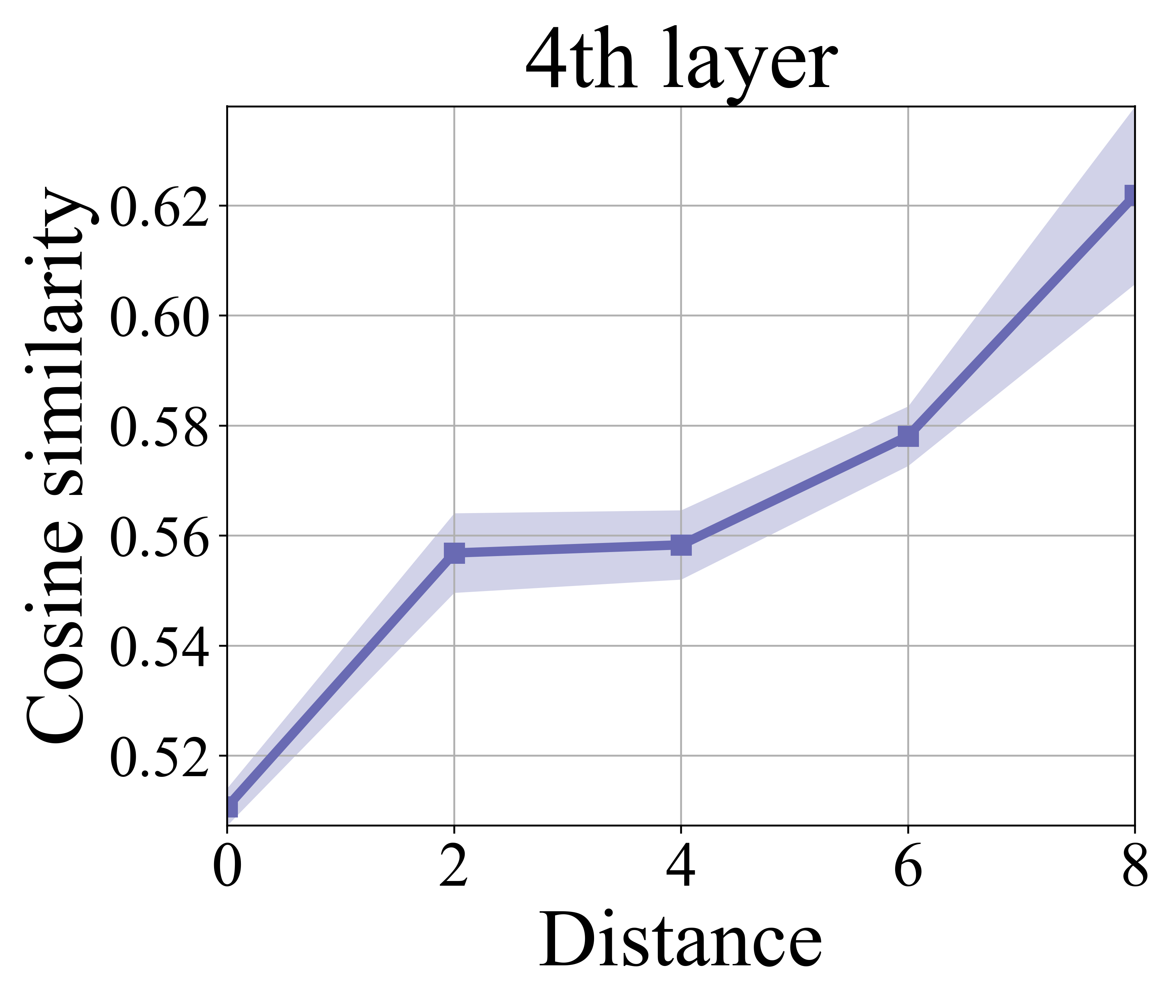}
        \label{fig:scale3_cossim}
    \end{subfigure}
    \caption{\rvc{Average} cosine similarity between error pixels and representative anchors in each layer was computed based on \rvc{the} distance from incorrect prediction boundaries. The results demonstrate that samples located along the incorrect prediction boundaries are harder-negative samples compared with features in the inner region.}
    \label{fig:cossim_dist}
\end{figure}

Second, we examined the cosine similarity between representative anchors and all the negative samples by considering their distances in the distance map, \rvc{i.e.}~$\mathbf{D}^n_i$ in \cref{fig:ext_sampling}(a), for each layer. \rvc{The} lower distance \rvc{implies} that the negative samples are more likely to be from the edge regions of the wrongly predicted segments. A lower cosine similarity means that the vector that \rvc{is supposed to} be close to the anchor is far apart, implying that the negative samples are more challenging to discriminate in the feature space. 
As presented in \cref{fig:cossim_dist}, the features existing along the boundaries of the incorrect prediction regions, which have lower distance values, are harder to discriminate well in all encoder layers. Thus, it corroborates that our BANE sampling successfully prioritizes harder-negative samples, triggering more desirable gradient contributions for our contextual contrastive learning. 


As a result, these analyses support our key claim that our multi-scale-aware representative anchors align features well on the embedding vector space and our BANE sampling successfully chooses informative negative examples.



\begin{table}[t]
\scriptsize
\centering
\renewcommand{\arraystretch}{1.3}
\begin{tabular}{c|l|cccc}  
\hline
\rowcolor[HTML]{DAE8FC}
 & \multicolumn{1}{c|}{\begin{tabular}[c]{@{}c@{}}Method\end{tabular}}                                         & A $\downarrow$                                & U $\uparrow$                               & U$_3$ $\uparrow$  & U$_5$ $\uparrow$                 \\ \hline
                                                      &                                          &                                          &                                          \\[-0.5ex]
& \multirow{-2}{*}{HRNet}                             & \multirow{-2}{*}{0.70}                  & \multirow{-2}{*}{0.98}                  & \multirow{-2}{*}{0.47} &\multirow{-2}{*}{0.55}                  \\[-1.0ex]
&                                                     & 0.53                                    & 1.00                          & \textbf{0.50}   & 0.58                             \\[-0.7ex]
& \multirow{-2}{*}{HRNet + \cite{pissas2022multi}}        & {\color[HTML]{2D8C00} (-0.17)}          & {\color[HTML]{2D8C00} (+0.02)} & {\color[HTML]{2D8C00} \textbf{(+0.03)}} & {\color[HTML]{2D8C00} (+0.03)}  \\
& \cellcolor[HTML]{EFEFEF}                             & \cellcolor[HTML]{EFEFEF}\textbf{0.42}                           & \cellcolor[HTML]{EFEFEF}\textbf{1.01}                                    & \cellcolor[HTML]{EFEFEF}\textbf{0.50}   & \cellcolor[HTML]{EFEFEF}\textbf{0.59}                                 \\[-0.7ex]
\multirow{-5}{*}{\rotatebox{90}{Cityscapes}} & \multirow{-2}{*}{\cellcolor[HTML]{EFEFEF}{HRNet + Ours}}                    & \cellcolor[HTML]{EFEFEF}{\color[HTML]{2D8C00} \textbf{(-0.28)}} & \cellcolor[HTML]{EFEFEF}{\color[HTML]{2D8C00} \textbf{(+0.03)}}          & \cellcolor[HTML]{EFEFEF}{\color[HTML]{2D8C00} \textbf{(+0.03)}}  & \cellcolor[HTML]{EFEFEF}{\color[HTML]{2D8C00} \textbf{(+0.04)}}        \\ \hline  
&                                                    &                                          &                                          &                                          \\[-0.5ex]
& \multirow{-2}{*}{OCRNet}                                                & \multirow{-2}{*}{0.60}                  & \multirow{-2}{*}{1.09}                  & \multirow{-2}{*}{0.64} &\multirow{-2}{*}{0.72}                 \\[-1.0ex]
&                                       &0.55                                     & \textbf{1.10}                          & 0.67   &0.74                             \\[-0.7ex]
& \multirow{-2}{*}{OCRNet + \cite{pissas2022multi}}                     & {\color[HTML]{2D8C00} (-0.05)}          & {\color[HTML]{2D8C00} \textbf{(+0.01)}} & {\color[HTML]{2D8C00} (+0.03)} & {\color[HTML]{2D8C00} (+0.02)}  \\
& \cellcolor[HTML]{EFEFEF}                             & \cellcolor[HTML]{EFEFEF}\textbf{0.52}                           & \cellcolor[HTML]{EFEFEF}\textbf{1.10}                                & \cellcolor[HTML]{EFEFEF}\textbf{0.68}   & \cellcolor[HTML]{EFEFEF}\textbf{0.75}                                  \\[-0.7ex]
\multirow{-6}{*}{\vspace{-0.4cm}\rotatebox{90}{CamVid}} & \multirow{-2}{*}{\cellcolor[HTML]{EFEFEF}{OCRNet + Ours}}                   & {\cellcolor[HTML]{EFEFEF}\color[HTML]{2D8C00} \textbf{(-0.08)}} & {\cellcolor[HTML]{EFEFEF}\color[HTML]{2D8C00} \textbf{(+0.01)}}          & {\cellcolor[HTML]{EFEFEF}\color[HTML]{2D8C00} \textbf{(+0.04)}}  & {\cellcolor[HTML]{EFEFEF}\color[HTML]{2D8C00} \textbf{(+0.03)}}     \\ \hline
\end{tabular}
\caption{Ablation analysis of Alignment (A), Uniformity (U), and the $l$-closest Neighborhood Uniformity ($\text{U}_l$) on the Cityscapes and CamVid datasets with HRNet~\cite{sun2019high} and OCRNet~\cite{chen2017rethinking} as segmentation models.}
\label{tab:alginment}
\end{table}


\newcolumntype{L}[1]{>{\raggedright\let\newline\\\arraybackslash\hspace{0pt}}m{#1}}
\newcolumntype{C}[1]{>{\centering\let\newline\\\arraybackslash\hspace{0pt}}m{#1}}
\newcolumntype{R}[1]{>{\raggedleft\let\newline\\\arraybackslash\hspace{0pt}}m{#1}}

\subsection{Ablation study}

\noindent\textbf{The impact of individual component\rvc{.}} To further examine the effectiveness
of each module more closely, we conducted an ablation study, as shown in \rvc{\Cref{tab:method_abl}}.
Applying our contextual contrastive learning led to enhancements of mIoU.
In particular, compared with \rvvc{ContrastiveSeg~\cite{wang2021exploring}}, the combination of our methods exhibits a larger gap in performance increase, which supports our key claim that the negative samples chosen by our BANE sampling on the contrastive learning is helpful by making vectors on the embedded space more distinct.
These results highlight the effectiveness of our approach, demonstrating a substantial increase in mIoU, making it a promising solution for semantic segmentation tasks across diverse datasets.

\begin{table}[t]
\scriptsize
\centering
\renewcommand{\arraystretch}{1.5}
\begin{tabular}{C{0.3cm} | C{1.3cm} C{1.5cm} C{1.5cm} C{1.5cm}} 
\hline
\rowcolor[HTML]{DAE8FC}
&                                   & \multicolumn{2}{c}{Sampling}                                 &                                                \\ \cline{3-4}
\rowcolor[HTML]{DAE8FC}
\multirow{-2}{*}{}                   & \multirow{-2}{*}{\rvvc{CCL (Ours)}}             & Semi-hard~\cite{wang2021exploring}                    & BANE~(Ours)                               & \multirow{-2}{*}{mIOU (\%)}                    \\ \hline
                                                          &                                   &                          &                                   & 78.48                                          \\
                                                         & \checkmark                         &                          &                                   & 81.88 {\color[HTML]{2D8C00} (+3.40)}                                  \\
                                                         & \checkmark                         & \checkmark                &                                   & 82.01 {\color[HTML]{2D8C00} (+3.53)}                                  \\
\multirow{-4}{*}{\vspace{-0.1cm}\rotatebox{90}{Cityscapes}}    & \cellcolor[HTML]{EFEFEF}\checkmark & \cellcolor[HTML]{EFEFEF} & \cellcolor[HTML]{EFEFEF}\checkmark & \cellcolor[HTML]{EFEFEF}\textbf{82.20{\color[HTML]{2D8C00}~(+3.72)}} \\ \hline
                                                         &                                   &                          &                                   & 82.17                                          \\
                                                         & \checkmark                         &                          &                                   & 83.14 {\color[HTML]{2D8C00} (+0.97)}                                  \\
                                                         & \checkmark                         & \checkmark                &                                   & 83.38 {\color[HTML]{2D8C00}(+1.21)}                                           \\
\multirow{-4}{*}{\vspace{-0.1cm}\rotatebox{90} {CamVid}}      & \cellcolor[HTML]{EFEFEF}\checkmark & \cellcolor[HTML]{EFEFEF} & \cellcolor[HTML]{EFEFEF}\checkmark & \cellcolor[HTML]{EFEFEF}\textbf{84.33{\color[HTML]{2D8C00}~(+2.16)}} \\ \hline
\end{tabular}
\caption{Ablation study: performance according to the presence or
absence of each component of our proposed method on the Cityscapes and CamVid datasets with HRNet~\cite{sun2019high} (\rvvc{CCL}: contextual contrastive learning).}
\label{tab:method_abl}
\end{table}

\noindent\textbf{Analysis of anchor fusion weight and ratio of boundary-aware negative sampling\rvc{.}}
\rvc{\Cref{tab:w_h}} demonstrates the proposed method's resilience to weight selection variations. The sum of $w_h$ and $w_l$ is 1. Except when $w_h$ is set to 0 \rvc{or} 1, the proposed method consistently improves the performance. Thus, \rvc{\Cref{tab:w_h}} shows that our proposed method is stable with regard to hyperparameter tuning.
In addition, we examine the impact of BANE sampling using different sampling ratios $K$, which is presented in~\cref{sec:boundary-aware}. As shown in\rvc{~\Cref{tab:ratio}}, our analysis reveals that the sampling method mostly enhances performance with \rvc{the ratio of} 50\%. However, the performance declines when excessive negative sampling is applied because it leads to local minima~\cite{schroff2015facenet, xie2022delving,cai2020all}. \abc{More ablation studies on hyper-parameters are presented in the supplementary material.}

\begin{table}[t]
\scriptsize
\centering
\renewcommand{\arraystretch}{1.3}
\begin{tabular}{C{1.2cm} | C{0.6cm} C{0.6cm} C{0.6cm} C{0.6cm} C{0.6cm} C{0.6cm}}
\hline
\rowcolor[HTML]{DAE8FC}
         $w_{h}$ & 0.0 & 0.3 & 0.5 & 0.7 & 1.0 \\ \hline
mIoU (\%) & 81.15 & 81.27 & 81.80 & \textbf{81.88} & 81.31 \\ \hline
\end{tabular}
\caption{Comparison of different weights \rvc{for} the representative anchor fusion on \rvc{Cityscapes-\texttt{val}} with HRNet~\cite{sun2019high}. The sum of $w_h$ and $w_l$ is equal to 1.}
\label{tab:w_h}
\end{table}

\begin{table}[t]
\scriptsize
\centering
\renewcommand{\arraystretch}{1.3}
\begin{tabular}{C{1.4cm} | C{0.6cm} C{0.6cm} C{0.6cm} C{0.6cm} C{0.6cm} C{0.6cm}}
\hline
\rowcolor[HTML]{DAE8FC}
        Ratio $K$  (\%)& 0 & 10 & 30 & 50        & 70 & 100 \\ \hline
mIoU (\%) & 81.88 & 81.89   & 81.58   & \textbf{82.20} & 81.58   & 81.10 \\ \hline
\end{tabular}
\caption{Comparison of different sampling ratios \rvc{for} boundary-aware negative sampling on \rvc{Cityscapes-\texttt{val} with HRNet~\cite{sun2019high}.}}
\label{tab:ratio}
\end{table}

\subsection{Conclusions}
In this paper, we have proposed a novel boundary-aware contrastive learning for semantic segmentation, called \textit{Contextrast}. By leveraging multi-scale contextual contrastive learning, we enable the network capture local/global context information and consistently understand their relationship. In particular, we demonstrate our BANE sampling substantially increases mIoU by providing more harder negative samples on the contrastive learning stage. Consequently, our approach achieved promising results compared with other contrastive learning approaches on public datasets. 


\noindent{\textbf{Acknowledgements.}}
This research was supported in part by the KAIST Convergence Research Institute Operation Program and in part by Korea Evaluation Institute Of Industrial Technology (KEIT) grant funded by the Korea government(MOTIE) (No.20023455, Development of Cooperate Mapping, Environment Recognition and Autonomous Driving Technology for Multi Mobile Robots Operating in Large-scale Indoor Workspace). We thank KI Cloud of Division of National Supercomputing Center, Korea Institute of Science and Technology Information(KISTI).

{
    \small
    \bibliographystyle{ieeenat_fullname}
    \bibliography{main}

\begin{thebibliography}{56}
\providecommand{\natexlab}[1]{#1}
\providecommand{\url}[1]{\texttt{#1}}
\expandafter\ifx\csname urlstyle\endcsname\relax
  \providecommand{\doi}[1]{doi: #1}\else
  \providecommand{\doi}{doi: \begingroup \urlstyle{rm}\Url}\fi

\bibitem[Brostow et~al.(2009)Brostow, Fauqueur, and Cipolla]{brostow2009semantic}
Gabriel~J Brostow, Julien Fauqueur, and Roberto Cipolla.
\newblock Semantic object classes in video: A high-definition ground truth database.
\newblock \emph{Pattern Recognition Letters}, 30\penalty0 (2):\penalty0 88--97, 2009.

\bibitem[Caesar et~al.(2018)Caesar, Uijlings, and Ferrari]{caesar2018coco}
Holger Caesar, Jasper Uijlings, and Vittorio Ferrari.
\newblock {COCO-stuff: Thing and stuff classes in context}.
\newblock In \emph{Proceedings of the IEEE/CVF Conference on Computer Vision and Pattern Recognition}, pages 1209--1218, 2018.

\bibitem[Cai et~al.(2020)Cai, Frankle, Schwab, and Morcos]{cai2020all}
Tiffany~Tianhui Cai, Jonathan Frankle, David~J Schwab, and Ari~S Morcos.
\newblock Are all negatives created equal in contrastive instance discrimination?
\newblock \emph{arXiv preprint arXiv:2010.06682}, 2020.

\bibitem[Chen et~al.(2014)Chen, Papandreou, Kokkinos, Murphy, and Yuille]{chen2014semantic}
Liang-Chieh Chen, George Papandreou, Iasonas Kokkinos, Kevin Murphy, and Alan~L Yuille.
\newblock {Semantic image segmentation with deep convolutional nets and fully connected CRFs}.
\newblock \emph{arXiv preprint arXiv:1412.7062}, 2014.

\bibitem[Chen et~al.(2017{\natexlab{a}})Chen, Papandreou, Kokkinos, Murphy, and Yuille]{chen2017deeplab}
Liang-Chieh Chen, George Papandreou, Iasonas Kokkinos, Kevin Murphy, and Alan~L Yuille.
\newblock {DeepLab: Semantic image segmentation with deep convolutional nets, atrous convolution, and fully connected CRFs}.
\newblock \emph{IEEE Transactions on Pattern Analysis and Machine Intelligence}, 40\penalty0 (4):\penalty0 834--848, 2017{\natexlab{a}}.

\bibitem[Chen et~al.(2017{\natexlab{b}})Chen, Papandreou, Schroff, and Adam]{chen2017rethinking}
Liang-Chieh Chen, George Papandreou, Florian Schroff, and Hartwig Adam.
\newblock Rethinking atrous convolution for semantic image segmentation.
\newblock \emph{arXiv preprint arXiv:1706.05587}, 2017{\natexlab{b}}.

\bibitem[Chen et~al.(2018)Chen, Zhu, Papandreou, Schroff, and Adam]{chen2018encoder}
Liang-Chieh Chen, Yukun Zhu, George Papandreou, Florian Schroff, and Hartwig Adam.
\newblock Encoder-decoder with atrous separable convolution for semantic image segmentation.
\newblock In \emph{Proceedings of the European Conference on Computer Vision}, pages 801--818, 2018.

\bibitem[Choi et~al.(2020)Choi, Kim, and Choo]{choi2020cars}
Sungha Choi, Joanne~T Kim, and Jaegul Choo.
\newblock {Cars can't fly up in the sky: Improving urban-scene segmentation via height-driven attention networks}.
\newblock In \emph{Proceedings of the IEEE/CVF Conference on Computer Vision and Pattern Recognitionn}, pages 9373--9383, 2020.

\bibitem[Cordts et~al.(2016)Cordts, Omran, Ramos, Rehfeld, Enzweiler, Benenson, Franke, Roth, and Schiele]{cordts2016cityscapes}
Marius Cordts, Mohamed Omran, Sebastian Ramos, Timo Rehfeld, Markus Enzweiler, Rodrigo Benenson, Uwe Franke, Stefan Roth, and Bernt Schiele.
\newblock The {C}ityscapes dataset for semantic urban scene understanding.
\newblock In \emph{Proceedings of the IEEE/CVF Conference on Computer Vision and Pattern Recognition}, pages 3213--3223, 2016.

\bibitem[Deng et~al.(2009)Deng, Dong, Socher, Li, Li, and Fei-Fei]{deng2009imagenet}
Jia Deng, Wei Dong, Richard Socher, Li-Jia Li, Kai Li, and Li Fei-Fei.
\newblock {ImageNet: A large-scale hierarchical image database}.
\newblock In \emph{Proceedings of the IEEE/CVF Conference on Computer Vision and Pattern Recognition}, pages 248--255, 2009.

\bibitem[Ding et~al.(2019)Ding, Jiang, Shuai, Liu, and Wang]{ding2019semantic}
Henghui Ding, Xudong Jiang, Bing Shuai, Ai~Qun Liu, and Gang Wang.
\newblock Semantic correlation promoted shape-variant context for segmentation.
\newblock In \emph{Proceedings of the IEEE/CVF Conference on Computer Vision and Pattern Recognition}, pages 8885--8894, 2019.

\bibitem[Duc et~al.(2022)Duc, Oanh, Thuy, Triet, and Dinh]{duc2022colonformer}
Nguyen~Thanh Duc, Nguyen~Thi Oanh, Nguyen~Thi Thuy, Tran~Minh Triet, and Viet~Sang Dinh.
\newblock {ColonFormer: An efficient transformer-based method for colon polyp segmentation}.
\newblock \emph{IEEE Access}, 10:\penalty0 80575--80586, 2022.

\bibitem[Dumitru et~al.(2023)Dumitru, Peteleaza, and Craciun]{dumitru2023using}
Razvan-Gabriel Dumitru, Darius Peteleaza, and Catalin Craciun.
\newblock {Using DUCK-Net for polyp image segmentation}.
\newblock \emph{Scientific Reports}, 13\penalty0 (1):\penalty0 9803, 2023.

\bibitem[Fu et~al.(2019)Fu, Liu, Tian, Li, Bao, Fang, and Lu]{fu2019dual}
Jun Fu, Jing Liu, Haijie Tian, Yong Li, Yongjun Bao, Zhiwei Fang, and Hanqing Lu.
\newblock Dual attention network for scene segmentation.
\newblock In \emph{Proceedings of the IEEE/CVF Conference on Computer Vision and Pattern Recognition}, pages 3146--3154, 2019.

\bibitem[Gutmann and Hyv{\"a}rinen(2010)]{gutmann2010noise}
Michael Gutmann and Aapo Hyv{\"a}rinen.
\newblock Noise-contrastive estimation: A new estimation principle for unnormalized statistical models.
\newblock In \emph{Proceedings of the International Conference on Artificial Intelligence and Statistics}, pages 297--304. JMLR, 2010.

\bibitem[Hong et~al.(2021)Hong, Pan, Sun, and Jia]{hong2021deep}
Yuanduo Hong, Huihui Pan, Weichao Sun, and Yisong Jia.
\newblock Deep dual-resolution networks for real-time and accurate semantic segmentation of road scenes.
\newblock \emph{arXiv preprint arXiv:2101.06085}, 2021.

\bibitem[Hu et~al.(2021)Hu, Cui, and Wang]{hu2021region}
Hanzhe Hu, Jinshi Cui, and Liwei Wang.
\newblock Region-aware contrastive learning for semantic segmentation.
\newblock In \emph{Proceedings of the IEEE/CVF International Conference on Computer Vision}, pages 16291--16301, 2021.

\bibitem[Hurtado and Valada(2022)]{hurtado2022semantic}
Juana~Valeria Hurtado and Abhinav Valada.
\newblock Semantic scene segmentation for robotics.
\newblock In \emph{Deep Learning for Robot Perception and Cognition}, pages 279--311. Elsevier, 2022.

\bibitem[Huynh et~al.(2021)Huynh, Tran, Luu, and Hoai]{huynh2021progressive}
Chuong Huynh, Anh~Tuan Tran, Khoa Luu, and Minh Hoai.
\newblock Progressive semantic segmentation.
\newblock In \emph{Proceedings of the {IEEE}/CVF Conference on Computer Vision and Pattern Recognition}, pages 16755--16764, 2021.

\bibitem[Kalantidis et~al.(2020)Kalantidis, Sariyildiz, Pion, Weinzaepfel, and Larlus]{kalantidis2020hard}
Yannis Kalantidis, Mert~Bulent Sariyildiz, Noe Pion, Philippe Weinzaepfel, and Diane Larlus.
\newblock Hard negative mixing for contrastive learning.
\newblock \emph{Advances in Neural Information Processing Systems}, 33:\penalty0 21798--21809, 2020.

\bibitem[Ke et~al.(2018)Ke, Hwang, Liu, and Yu]{ke2018adaptive}
Tsung-Wei Ke, Jyh-Jing Hwang, Ziwei Liu, and Stella~X Yu.
\newblock Adaptive affinity fields for semantic segmentation.
\newblock In \emph{Proceedings of the European Conference on Computer Vision}, pages 587--602, 2018.

\bibitem[Khosla et~al.(2020)Khosla, Teterwak, Wang, Sarna, Tian, Isola, Maschinot, Liu, and Krishnan]{khosla2020supervised}
Prannay Khosla, Piotr Teterwak, Chen Wang, Aaron Sarna, Yonglong Tian, Phillip Isola, Aaron Maschinot, Ce Liu, and Dilip Krishnan.
\newblock Supervised contrastive learning.
\newblock \emph{Advances in Neural Information Processing Systems}, 33:\penalty0 18661--18673, 2020.

\bibitem[Kimmel et~al.(1996)Kimmel, Kiryati, and Bruckstein]{kimmel1996sub}
Ron Kimmel, Nahum Kiryati, and Alfred~M Bruckstein.
\newblock Sub-pixel distance maps and weighted distance transforms.
\newblock \emph{Journal of Mathematical Imaging and Vision}, 6:\penalty0 223--233, 1996.

\bibitem[Li et~al.(2018)Li, Xiong, An, and Wang]{li2018pyramid}
Hanchao Li, Pengfei Xiong, Jie An, and Lingxue Wang.
\newblock Pyramid attention network for semantic segmentation.
\newblock \emph{arXiv preprint arXiv:1805.10180}, 2018.

\bibitem[Li et~al.(2022{\natexlab{a}})Li, Zhou, Wang, Li, and Yang]{li2022deep}
Liulei Li, Tianfei Zhou, Wenguan Wang, Jianwu Li, and Yi Yang.
\newblock Deep hierarchical semantic segmentation.
\newblock In \emph{Proceedings of the IEEE/CVF Conference on Computer Vision and Pattern Recognition}, pages 1246--1257, 2022{\natexlab{a}}.

\bibitem[Li et~al.(2022{\natexlab{b}})Li, Cao, Yuan, Fan, Yang, Feris, Indyk, and Katabi]{li2022targeted}
Tianhong Li, Peng Cao, Yuan Yuan, Lijie Fan, Yuzhe Yang, Rogerio~S Feris, Piotr Indyk, and Dina Katabi.
\newblock Targeted supervised contrastive learning for long-tailed recognition.
\newblock In \emph{Proceedings of the IEEE/CVF Conference on Computer Vision and Pattern Recognition}, pages 6918--6928, 2022{\natexlab{b}}.

\bibitem[Liu et~al.(2021)Liu, Lin, Cao, Hu, Wei, Zhang, Lin, and Guo]{liu2021swin}
Ze Liu, Yutong Lin, Yue Cao, Han Hu, Yixuan Wei, Zheng Zhang, Stephen Lin, and Baining Guo.
\newblock {Swin Transformer: Hierarchical vision transformer using shifted windows}.
\newblock In \emph{Proceedings of the IEEE/CVF International Conference on Computer Vision}, pages 10012--10022, 2021.

\bibitem[Long et~al.(2015)Long, Shelhamer, and Darrell]{long2015fully}
Jonathan Long, Evan Shelhamer, and Trevor Darrell.
\newblock Fully convolutional networks for semantic segmentation.
\newblock In \emph{Proceedings of the IEEE/CVF Conference on Computer Vision and Pattern Recognition}, pages 3431--3440, 2015.

\bibitem[Mottaghi et~al.(2014)Mottaghi, Chen, Liu, Cho, Lee, Fidler, Urtasun, and Yuille]{mottaghi2014role}
Roozbeh Mottaghi, Xianjie Chen, Xiaobai Liu, Nam-Gyu Cho, Seong-Whan Lee, Sanja Fidler, Raquel Urtasun, and Alan Yuille.
\newblock The role of context for object detection and semantic segmentation in the wild.
\newblock In \emph{Proceedings of the IEEE/CVF Conference on Computer Vision and Pattern Recognition}, pages 891--898, 2014.

\bibitem[Oord et~al.(2018)Oord, Li, and Vinyals]{oord2018representation}
Aaron van~den Oord, Yazhe Li, and Oriol Vinyals.
\newblock Representation learning with contrastive predictive coding.
\newblock \emph{arXiv preprint arXiv:1807.03748}, 2018.

\bibitem[Pissas et~al.(2022)Pissas, Ravasio, Cruz, and Bergeles]{pissas2022multi}
Theodoros Pissas, Claudio~S Ravasio, Lyndon~Da Cruz, and Christos Bergeles.
\newblock Multi-scale and cross-scale contrastive learning for semantic segmentation.
\newblock In \emph{Proceedings of the European Conference on Computer Vision}, pages 413--429. Springer, 2022.

\bibitem[Sanderson and Matuszewski(2022)]{sanderson2022fcn}
Edward Sanderson and Bogdan~J Matuszewski.
\newblock {FCN-transformer feature fusion for polyp segmentation}.
\newblock In \emph{Proceedings of the Annual Conference on Medical Image Understanding and Analysis}, pages 892--907. Springer, 2022.

\bibitem[Schroff et~al.(2015)Schroff, Kalenichenko, and Philbin]{schroff2015facenet}
Florian Schroff, Dmitry Kalenichenko, and James Philbin.
\newblock {FaceNet: A unified embedding for face recognition and clustering}.
\newblock In \emph{Proceedings of the IEEE/CVF Conference on Computer Vision and Pattern Recognition}, pages 815--823, 2015.

\bibitem[Strudel et~al.(2021)Strudel, Garcia, Laptev, and Schmid]{strudel2021segmenter}
Robin Strudel, Ricardo Garcia, Ivan Laptev, and Cordelia Schmid.
\newblock Segmenter: Transformer for semantic segmentation.
\newblock In \emph{Proceedings of the IEEE/CVF International Conference on Computer Vision}, pages 7262--7272, 2021.

\bibitem[Sun et~al.(2019)Sun, Zhao, Jiang, Cheng, Xiao, Liu, Mu, Wang, Liu, and Wang]{sun2019high}
Ke Sun, Yang Zhao, Borui Jiang, Tianheng Cheng, Bin Xiao, Dong Liu, Yadong Mu, Xinggang Wang, Wenyu Liu, and Jingdong Wang.
\newblock High-resolution representations for labeling pixels and regions.
\newblock \emph{arXiv preprint arXiv:1904.04514}, 2019.

\bibitem[Tan et~al.(2022)Tan, Wu, and Pi]{tan2022semantic}
Haoru Tan, Sitong Wu, and Jimin Pi.
\newblock Semantic diffusion network for semantic segmentation.
\newblock \emph{Advances in Neural Information Processing Systems}, 35:\penalty0 8702--8716, 2022.

\bibitem[Tzelepi and Tefas(2021)]{tzelepi2021semantic}
Maria Tzelepi and Anastasios Tefas.
\newblock Semantic scene segmentation for robotics applications.
\newblock In \emph{Proceedings of the International Conference on Information, Intelligence, Systems \& Applications}, pages 1--4, 2021.

\bibitem[Wang et~al.(2022)Wang, Zhang, Cui, Ren, Yang, Xie, Hua, Bao, and Xu]{wang2022active}
Chi Wang, Yunke Zhang, Miaomiao Cui, Peiran Ren, Yin Yang, Xuansong Xie, Xian-Sheng Hua, Hujun Bao, and Weiwei Xu.
\newblock Active boundary loss for semantic segmentation.
\newblock In \emph{Proceedings of the AAAI Conference on Artificial Intelligence}, pages 2397--2405, 2022.

\bibitem[Wang et~al.(2020)Wang, Sun, Cheng, Jiang, Deng, Zhao, Liu, Mu, Tan, Wang, et~al.]{wang2020deep}
Jingdong Wang, Ke Sun, Tianheng Cheng, Borui Jiang, Chaorui Deng, Yang Zhao, Dong Liu, Yadong Mu, Mingkui Tan, Xinggang Wang, et~al.
\newblock Deep high-resolution representation learning for visual recognition.
\newblock \emph{IEEE Transactions on Pattern Analysis and Machine Intelligence}, 43\penalty0 (10):\penalty0 3349--3364, 2020.

\bibitem[Wang et~al.(2023{\natexlab{a}})Wang, Huang, Tang, Meng, Su, and Song]{wang2203stepwise}
J Wang, Q Huang, F Tang, J Meng, J Su, and S Song.
\newblock Stepwise feature fusion: Local guides global.
\newblock \emph{arXiv preprint arXiv:2203.03635}, 2023{\natexlab{a}}.

\bibitem[Wang et~al.(2021)Wang, Zhou, Yu, Dai, Konukoglu, and Van~Gool]{wang2021exploring}
Wenguan Wang, Tianfei Zhou, Fisher Yu, Jifeng Dai, Ender Konukoglu, and Luc Van~Gool.
\newblock Exploring cross-image pixel contrast for semantic segmentation.
\newblock In \emph{Proceedings of the IEEE/CVF International Conference on Computer Vision}, pages 7303--7313, 2021.

\bibitem[Wang et~al.(2023{\natexlab{b}})Wang, Dai, Chen, Huang, Li, Zhu, Hu, Lu, Lu, Li, et~al.]{wang2023internimage}
Wenhai Wang, Jifeng Dai, Zhe Chen, Zhenhang Huang, Zhiqi Li, Xizhou Zhu, Xiaowei Hu, Tong Lu, Lewei Lu, Hongsheng Li, et~al.
\newblock {Internimage: Exploring large-scale vision foundation models with deformable convolutions}.
\newblock In \emph{Proceedings of the IEEE/CVF Conference on Computer Vision and Pattern Recognition}, pages 14408--14419, 2023{\natexlab{b}}.

\bibitem[Woo et~al.(2023)Woo, Debnath, Hu, Chen, Liu, Kweon, and Xie]{woo2023convnext}
Sanghyun Woo, Shoubhik Debnath, Ronghang Hu, Xinlei Chen, Zhuang Liu, In~So Kweon, and Saining Xie.
\newblock {ConvNext v2: Co-designing and scaling ConvNets with masked autoencoders}.
\newblock In \emph{Proceedings of the IEEE/CVF Conference on Computer Vision and Pattern Recognition}, pages 16133--16142, 2023.

\bibitem[Xiao et~al.(2018)Xiao, Liu, Zhou, Jiang, and Sun]{xiao2018unified}
Tete Xiao, Yingcheng Liu, Bolei Zhou, Yuning Jiang, and Jian Sun.
\newblock Unified perceptual parsing for scene understanding.
\newblock In \emph{Proceedings of the European Conference on Computer Vision}, pages 418--434, 2018.

\bibitem[Xie et~al.(2022)Xie, Zhan, Liu, Ong, and Loy]{xie2022delving}
Jiahao Xie, Xiaohang Zhan, Ziwei Liu, Yew-Soon Ong, and Chen~Change Loy.
\newblock Delving into inter-image invariance for unsupervised visual representations.
\newblock \emph{International Journal of Computer Vision}, 130\penalty0 (12):\penalty0 2994--3013, 2022.

\bibitem[Xu et~al.(2023)Xu, Xiong, and Bhattacharyya]{xu2023pidnet}
Jiacong Xu, Zixiang Xiong, and Shankar~P Bhattacharyya.
\newblock {PIDNet: A real-time semantic segmentation network inspired by PID controllers}.
\newblock In \emph{Proceedings of the IEEE/CVF Conference on Computer Vision and Pattern Recognition}, pages 19529--19539, 2023.

\bibitem[Yu et~al.(2018)Yu, Wang, Peng, Gao, Yu, and Sang]{yu2018learning}
Changqian Yu, Jingbo Wang, Chao Peng, Changxin Gao, Gang Yu, and Nong Sang.
\newblock Learning a discriminative feature network for semantic segmentation.
\newblock In \emph{Proceedings of the IEEE/CVF Conference on Computer Vision and Pattern Recognition}, pages 1857--1866, 2018.

\bibitem[Yu et~al.(2020)Yu, Wang, Gao, Yu, Shen, and Sang]{yu2020context}
Changqian Yu, Jingbo Wang, Changxin Gao, Gang Yu, Chunhua Shen, and Nong Sang.
\newblock Context prior for scene segmentation.
\newblock In \emph{Proceedings of the IEEE/CVF Conference on Computer Vision and Pattern Recognition}, pages 12416--12425, 2020.

\bibitem[Yuan et~al.(2020{\natexlab{a}})Yuan, Chen, and Wang]{yuan2020object}
Yuhui Yuan, Xilin Chen, and Jingdong Wang.
\newblock Object-contextual representations for semantic segmentation.
\newblock In \emph{Proceedings of the European Conference on Computer Vision}, pages 173--190. Springer, 2020{\natexlab{a}}.

\bibitem[Yuan et~al.(2020{\natexlab{b}})Yuan, Xie, Chen, and Wang]{yuan2020segfix}
Yuhui Yuan, Jingyi Xie, Xilin Chen, and Jingdong Wang.
\newblock {SegFix: Model-agnostic boundary refinement for segmentation}.
\newblock In \emph{Proceedings of the European Conference on Computer Vision}, pages 489--506. Springer, 2020{\natexlab{b}}.

\bibitem[Yurtkulu et~al.(2019)Yurtkulu, {\c{S}}ahin, and Unal]{yurtkulu2019semantic}
Salih~Can Yurtkulu, Yusuf~H{\"u}seyin {\c{S}}ahin, and Gozde Unal.
\newblock Semantic segmentation with extended deeplabv3 architecture.
\newblock In \emph{Proceedings of the Signal Processing and Communications Applications Conference}, pages 1--4. IEEE, 2019.

\bibitem[Zhang et~al.(2019)Zhang, Li, Arnab, Yang, Tong, and Torr]{zhang2019dual}
Li Zhang, Xiangtai Li, Anurag Arnab, Kuiyuan Yang, Yunhai Tong, and Philip~HS Torr.
\newblock Dual graph convolutional network for semantic segmentation.
\newblock \emph{arXiv preprint arXiv:1909.06121}, 2019.

\bibitem[Zhao et~al.(2017)Zhao, Shi, Qi, Wang, and Jia]{zhao2017pyramid}
Hengshuang Zhao, Jianping Shi, Xiaojuan Qi, Xiaogang Wang, and Jiaya Jia.
\newblock Pyramid scene parsing network.
\newblock In \emph{Proceedings of the IEEE/CVF Conference on Computer Vision and Pattern Recognition}, pages 2881--2890, 2017.

\bibitem[Zhong et~al.(2023)Zhong, Cui, Yang, Wu, Qi, Zhang, and Jia]{zhong2023understanding}
Zhisheng Zhong, Jiequan Cui, Yibo Yang, Xiaoyang Wu, Xiaojuan Qi, Xiangyu Zhang, and Jiaya Jia.
\newblock Understanding imbalanced semantic segmentation through neural collapse.
\newblock In \emph{Proceedings of the IEEE/CVF Conference on Computer Vision and Pattern Recognition}, pages 19550--19560, 2023.

\bibitem[Zhou et~al.(2017)Zhou, Zhao, Puig, Fidler, Barriuso, and Torralba]{zhou2017scene}
Bolei Zhou, Hang Zhao, Xavier Puig, Sanja Fidler, Adela Barriuso, and Antonio Torralba.
\newblock {Scene parsing through ADE20K dataset}.
\newblock In \emph{Proceedings of the IEEE/CVF Conference on Computer Vision and Pattern Recognition}, pages 633--641, 2017.

\bibitem[Zhou et~al.(2019)Zhou, Hao, Zhang, Zou, and Zhang]{zhou2019fusion}
Jingchun Zhou, Mingliang Hao, Dehuan Zhang, Peiyu Zou, and Weishi Zhang.
\newblock {Fusion PSPNet image segmentation based method for multi-focus image fusion}.
\newblock \emph{IEEE Photonics Journal}, 11\penalty0 (6):\penalty0 1--12, 2019.

\end{thebibliography}
}
\end{document}